\documentclass[10pt,letterpaper]{article}
\usepackage[top=0.85in,footskip=0.75in,marginparwidth=2in]{geometry}

\usepackage[utf8]{inputenc}

\usepackage{cite}

\usepackage{nameref,hyperref}

\usepackage[right]{lineno}

\usepackage{microtype}
\DisableLigatures[f]{encoding = *, family = * }

\raggedright
\usepackage{changepage}

\usepackage[aboveskip=1pt,labelfont=bf,labelsep=period,singlelinecheck=off]{caption}

\usepackage{mathtools}
\usepackage{bm}
\usepackage{amssymb}

\makeatletter
\renewcommand{\@biblabel}[1]{\quad#1.}
\makeatother

\usepackage{lastpage,fancyhdr,graphicx}
\usepackage{epstopdf}
\fancyheadoffset[L]{0.25in}
\fancyfootoffset[L]{0.25in}

\usepackage{color}

\definecolor{Gray}{gray}{.25}

\usepackage{graphicx}

\usepackage{ragged2e}

\begin{document}
\vspace*{0.35in}

\begin{flushleft}
{\Large
\textbf\newline{Abdominal Multi-organ Segmentation with Organ-Attention Networks and Statistical Fusion}
}
\newline
\\
Yan Wang\textsuperscript{a,}\footnotemark[1],
Yuyin Zhou\textsuperscript{a,}\footnotemark[1],
Wei Shen\textsuperscript{b,a},
Seyoun Park\textsuperscript{c,*},
Elliot K. Fishman\textsuperscript{c},
Alan L. Yuille\textsuperscript{d,a},
\\
\bigskip
\textsuperscript{a} Department of Computer Science, Johns Hopkins University, USA
\\
\textsuperscript{b} Key Laboratory of Specialty Fiber Optics and Optical Access Networks, Shanghai University, China
\\
\textsuperscript{c} Department of Radiology and Radiological Science, Johns Hopkins University, USA
\\
\textsuperscript{d} Department of Cognitive Science, Johns Hopkins University, USA
\\
\bigskip
*spark139@jhmi.edu
\footnotetext[1]{The first two authors equally contributed to the work.}

\end{flushleft}

\justify
\section*{Abstract}
Accurate and robust segmentation of abdominal organs on CT is essential for many clinical applications such as computer-aided diagnosis and computer-aided surgery. But this task is challenging due to the weak boundaries of organs, the complexity of the background, and the variable sizes of different organs. To address these challenges, we introduce a novel framework for multi-organ segmentation of abdominal regions by using organ-attention networks with reverse connections (OAN-RCs) which are applied to 2D views, of the 3D CT volume, and output estimates which are combined by statistical fusion exploiting structural similarity. More specifically, OAN is a two-stage deep convolutional network, where deep network features from the first stage are combined with the original image, in a second stage, to reduce the complex background and enhance the discriminative information for the target organs. Intuitively, OAN reduces the effect of the complex background by focusing attention so that each organ only needs to be discriminated from its local background. RCs are added to the first stage to give the lower layers more semantic information thereby enabling them to adapt to the sizes of different organs. Our networks are trained on 2D views (slices) enabling us to use holistic information and allowing efficient computation (compared to using 3D patches). To compensate for the limited cross-sectional information of the original 3D volumetric CT, e.g., the connectivity between neighbor slices, multi-sectional images are reconstructed from the three different 2D view directions. Then we combine the segmentation results from the different views using statistical fusion, with a novel term relating the structural similarity of the 2D views to the original 3D structure. To train the network and evaluate results, $13$ structures were manually annotated by four human raters and confirmed by a senior expert on $236$ normal cases. We tested our algorithm by 4-fold cross-validation and computed Dice-S{\o}rensen similarity coefficients (DSC) and surface distances for evaluating our estimates of the $13$ structures. Our experiments show that the proposed approach gives strong results and outperforms 2D- and 3D-patch based state-of-the-art methods in terms of DSC and mean surface distances.


\section{Introduction}
\label{sec:Introduction}

Segmentation of the internal structures, like body organs, in medical images is an essential task for many clinical applications such as computer-aided diagnosis (CAD), computer-aided surgery (CAS) and radiation therapy  (RT). However, despite intensive studies of automatic or semi-automatic segmentation methods, there remain challenges which need to be overcome before these methods can be applied  to clinical environments. In particular, detailed abdominal organ segmentation on CT is a challenging task both for manual human annotation and for automatic segmentation algorithms for various reasons including the morphological complexity of the structures, the large variations between inter- and intra-subjects, and image characteristics such as low contrast of soft tissues.

Early studies of abdominal organ segmentation focused on specific single organs, for example relatively large isolated structures such as the liver \cite{Heimann2009,Mharib2012,Li2015} or critical structures such as blood vessels \cite{Kirbas2004,Lesage2009}. However, most of the algorithms were based on specific features of the target organ, and so extensibility to the simultaneous segmentation of multiple organs was limited. For multi-organ segmentation, atlas-based approaches were adopted for many applications \cite{Iglesias2015,Asman2013,Chu2013,Wolz2013,Kada2015,Zhuang2016,Karasawa2017}. The general framework of  atlas-based segmentations is to deformably register selected atlas images with segmented structures to the target image. Critical issues for this approach, which affect performance accuracy, include proper atlas selection, accurate deformable image registration, and label fusion. In particular,
for the abdominal region, inter-subject variations are relatively large compared with other parts of the body (e.g., the brain) so the segmentation results are dependent on deformable registration between inter-subjects from the limited set of atlases, which is a challenging problem that critically affects the final accuracies. In addition, computational time is strongly dependent on the number of atlases. Therefore, selection of the proper number and types of atlases is a critical factor for both of the accuracy and efficiency.

Recently, learning-based approaches exploiting large datasets have been applied to the segmentation of medical images \cite{Dou2016,Cicek2016,Milletari2016,Nascimento2016,Setio2016,Chen2017,Rajchl2017a,Havaei2017,Kamnitsas2017,Zu2017}. In particular, deep convolutional neural networks (CNN) have been very successful \cite{Dou2016,Cicek2016,Roth2015,Roth2016,Milletari2016,Setio2016,Chen2017,Rajchl2017a,Havaei2017,Kamnitsas2017}. Targets include regions in the brain \cite{Chen2017,Havaei2017,Kamnitsas2017}, chest \cite{Setio2016}, and abdomen \cite{Dou2016,Roth2015,Roth2016}. The performance results of CNNs for organs (and even tumors) reach, or outperform, alternative state-of-the-art methods. Unlike multi-atlas-based approaches, deep networks do not require selecting a specific atlas or require deformable registration from training sets to a target image. In this study, we apply deep network approaches to abdominal organ segmentation. 

Most studies based on deep networks, however,  focused on a single structure segmentation, particularly for abdominal regions, and there are few studies of  multi-organ segmentation partly due to technical challenges discussed later. We note that fully convolutional networks (FCNs) \cite{Long2015} have been generally accepted  for organ segmentations on CT scans \cite{Cicek2016,Zhou2017,Roth2017} partly because they give state-of-the-art performance for semantic segmentation of natural images \cite{Long2015, Chen2016}. But there are three major characteristics of abdominal CT which we must address in order to obtain strong performance on multi-organ segmentation.

Firstly, many abdominal organs have weak boundaries between spatially adjacent structures on CT, e.g. between the head of the pancreas and the duodenum.  In addition, the entire CT volume includes a large variety of different complex structures. Morphological and topological complexity includes anatomically connected structures such as the gastrointestinal (GI) track (stomach, duodenum, small bowel and colon) and vascular structures.  The correct anatomical borders between connected structures may not be always visible in CT, especially in sectional images (i.e., 2D slices), and may be indicated only by subtle texture and shape change, which causes uncertainty even for human experts. This makes it hard for deep networks to distinguish the target organs from the complex background.

Secondly, there are large variations in the relative sizes of different target organs, e.g. the liver compared to the gallbladder. This causes problems when applying deep networks to multi-organ segmentation
because lower layers typically lack semantic information when segmenting small structures. The same problem has been observed in semantic segmentation of natural images where the segmentation performance on  small regions is typically much worse than on large regions, motivating the need to introduce mechanisms which attend to the scale \cite{chen2016attention}. 

Thirdly, although CT scans are high-resolution three-dimensional volumes, most current deep network methods were designed for 2D images. To overcome the limitations of using 2D CNNs for 3D images, Setio \emph{et al.} \cite{Setio2016} used multiple 2D patches reconstructed from $9$ different directions around the target region for the task of pulmonary nodule detection. Zhuang \emph{et al.} \cite{Zhuang2016} used 2D axial, coronal, and sagittal slices for pancreas detection at the coarse level and also for segmentation at the finer level. More recently, there are studies which use 3D deep networks \cite{Cicek2016,Milletari2016,Roth2017,Kamnitsas2017,Roth2018}. These, however, are not networks that act on the entire 3D CT volume but instead are local patch-based approaches (due to complex challenges of 3D deep networks discussed later in this paragraph). To address the problems caused by restricting to image patches, \cite{Roth2017,Kamnitsas2017} used a hierarchical approach with multi-resolutions, which reduces the dimension of the whole volume for initial detection and focuses on smaller regions at the finer resolution. But this strategy is best suited to a single target structure. Roth \emph{et al.} \cite{Roth2018} applied a bigger patch size to deal with the whole dense pancreatic volume, but this was also for single pancreas segmentation and hard to extend to the whole abdominal region. In general, 3D deep networks face far greater complex challenges than 2D deep networks. Both approaches rely heavily on 
graphics processing units (GPUs) but these GPUs have limited memory size which makes it difficult 
when dealing with full 3D CT volumes compared to 2D CT slices (which require much less memory). In addition, 3D deep networks typically require many more parameters than 2D deep networks and hence require much more training data, unless they are restricted to patches. But there is limited training data for  abdominal CT images, because annotating them is challenging and requires expert human radiologists, which makes it particularly difficult to apply 3D deep networks to abdominal multi-organ segmentation. We have, however, implemented a 3D patch based approach for comparison. 

To deal with the technical difficulties for abdominal multi-organ segmentation on CT, we introduce a novel framework of an organ-attention 2D deep networks with reverse connections (OAN-RC) followed by 
statistical fusion to combine the information from the three different views exploiting structural similarity using local isotropic 3D patches. OAN is a two-stage deep network, which computes an organ-attention map (OAM) from typical probability map of labels for input images in the first stage and combines OAM to the original input image for the second stage. This two-stage strategy effectively reduces the complexity of the background while enhancing the discriminative information of target structures (by concentrating attention close to the target structures). By training OAM with additional deep network, uncertainties and errors from the first stage are adjusted and the fidelity of the final probability map is improved. In this procedure, we apply reverse connections \cite{Kong17} to the first stage so that we can localize organ information at different scales by assisting the lower layers with semantic information.

\begin{figure*}
  \centering
  \includegraphics[width=1.0\linewidth]{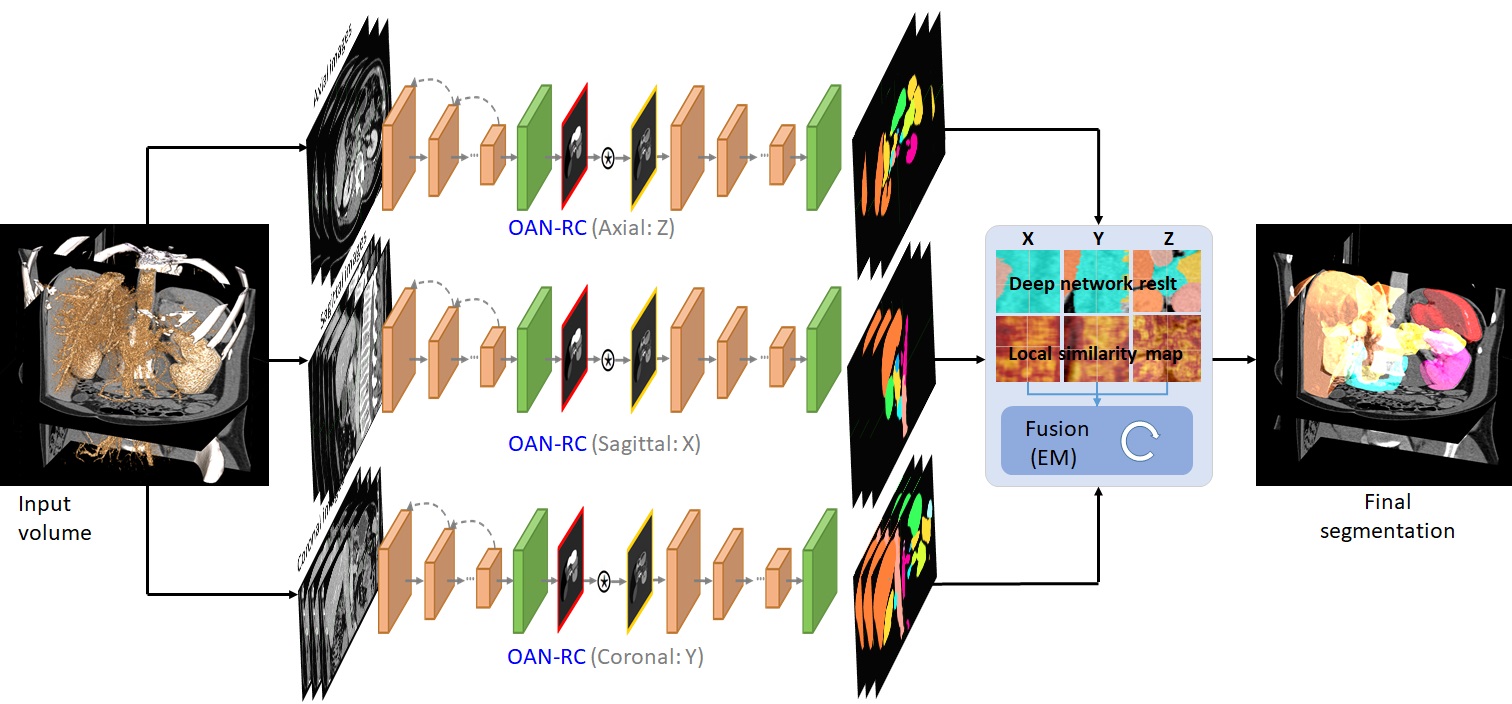}
  \caption{The overall framework.}
  \label{fig:OverallFramework}
\end{figure*}

More specifically, we apply OAN-RC to each sectional slice, which is an extreme form of anisotropic local patches but include the whole semantic (i.e. volume) information from one viewing direction. This yields  segmentation information from separate sets of multi-sectional images (axial, coronal, and sagittal planes in this study similarly to most of medical image platforms for 2D visualization). We statistically fuse the three sources of information using local isotropic 3D patches based on direction-dependent local structural similarity. The basic fusion framework uses expectation-maximization (EM) similar to \cite{Warfield2004,Asman2013}. But, unlike typical statistical fusion methods used for atlas-based segmentation, the input volumes and the target volumes for segmentation in our problem are the same. But different structures and texture patterns, from different viewing directions, will often generate nonidentical segmentations in 3D. Our strategy is to exploit structural similarity by computing a 
direction-dependent local property at each voxel. This models the structural similarity from the 2D images to the original 3D structure (in the 3D volume) by local weights. This structural statistical fusion improves our overall performance by combining the information from the three different views in a principled manner and also imposing local structure. 

Figure \ref{fig:OverallFramework} describes the graphical concept of our framework. Our proposed algorithm was tested on $236$ abdominal CT scans of normal cases collected as a part of FELIX project for pancreatic cancer research \cite{Lugo-Fagundo17}. By experiments, our method showed robust and high fidelities to the ground-truth for all target structures with smooth boundaries. It outperformed 3D patch-based algorithms as well as 2D-based in terms of DICE-similarity coefficient and average surface distance with memory and computational efficiency.

\section{Organ-Attention Networks with Reverse Connections}
\label{sec:OAN-RC}

Given a 3D volume of interest (VOI) of a scanned CT image $V \subset \mathbb{R}^3$, our goal is to find the label of each voxel $v \in V$. The target structures (i.e., the labeled structures) are restricted to be organs which do not overlap with each other, so every voxel $v$ should be assigned to a label in a finite set $\mathcal{L}$. In this section we introduce our proposed organ-attention networks with Reverse connections (annotated as OAN-RC) which is run separately on three different views, and then in the next section we describe our novel structural similarity statistical fusion method which combines the segmentation results obtained from the OAN-RCs on the three different views.
\subsection{Two-stage Organ Attention Network}
\label{sec:TwoStageOAN}

We first introduce the OAN, which is composed of two jointly optimized stages. The first stage (stage-I) transforms the organ segmentation probability map to provide spatial attention to the second stage (stage-II), so that the segmentation network trained in stage-II is more discriminative for segmenting organs (because it only has to deal with local context). To assist the lower layers in stage-I with more semantic information, we employ reverse connections (Sec. \ref{sec:ReverseConnection}), which pass semantic information down from high layers to low layers. The OAN is trained in an end-to-end fashion to enhance the learning ability of all stages.

The input images to our OAN are reconstructed 2D slices from axial, sagittal and coronal directions. Based on the normal vector directions of the sagittal ($X$), coronal ($Y$) and axial ($Z$) planes, we denote the 2D images by $\mathbf{I}_i^{X}$, $\mathbf{I}_j^Y$ and $\mathbf{I}_k^{Z}$ respectively, where $i=1,\ldots,n_{x},~j=1,\ldots,n_{y},~k=1,\ldots,n_{z}$ and $n_{x},~n_{y},~n_{z}$ are the numbers of slices for the three directions, respectively, and $\bigcup_{i} \mathbf{I}^{X} _{i} = \bigcup_{j} \mathbf{I}^{Y} _{j} = \bigcup_{k} \mathbf{I}^{Z} _{k} = V$. Following the work of \cite{Zhou2017}, we train an individual OAN for each direction.


\begin{figure}[ht]
\centering
\includegraphics[width=1.0\textwidth]{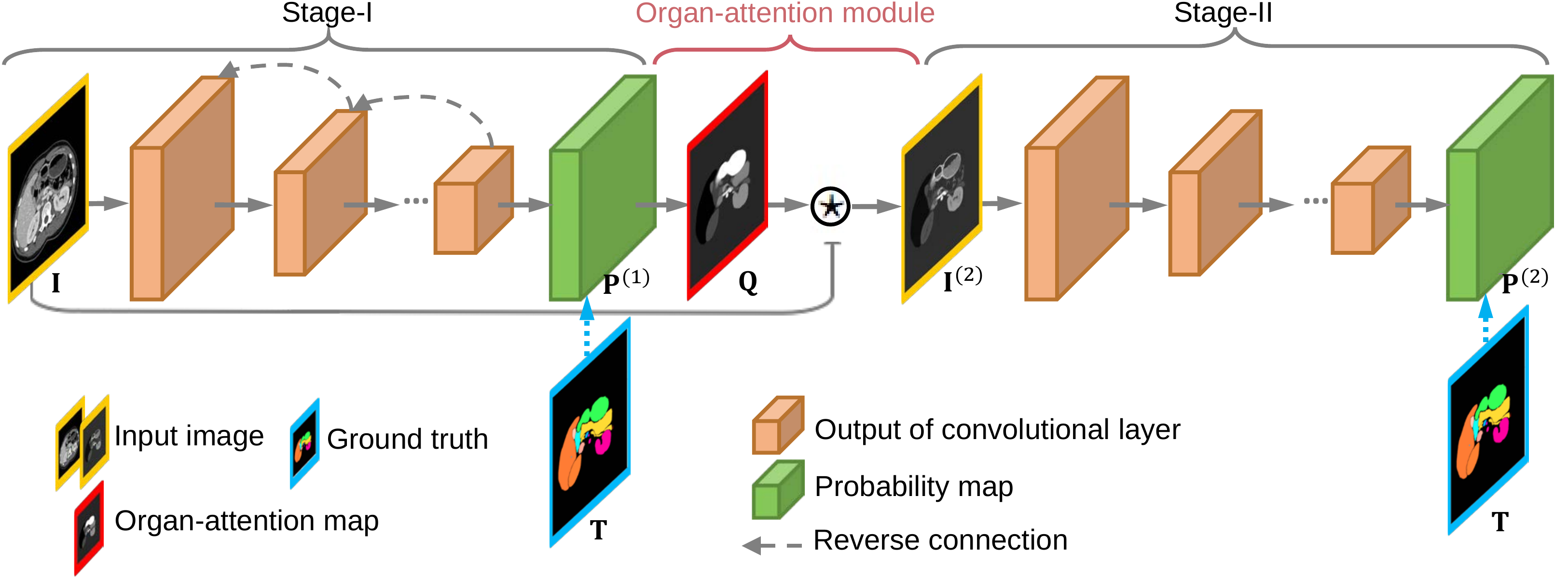}
\caption{The architecture of our two-stage organ-attention network with reverse connections. The organ-attention network (OAN) is composed of two jointly optimized stages, where the first stage (stage-I) transforms the organ segmentation probability map by spatial attention to the second stage (stage-II). Hence the organ segmentation map generated in the organ-attention module guides the latter computation. The reverse connections, described in Sec. \ref{sec:ReverseConnection}, modify the first stage of OAN as shown by dashed lines.}
\label{fig:2stageOAN}
\end{figure}

Fig.~\ref{fig:2stageOAN} illustrated our organ-attention-network architecture. The network consists of two stages, where each stage is a segmentation network. For notational simplicity, we denote an input 2D slice by $\mathbf{I}\subset \mathbb{R}^{H\times W}$ and its corresponding label map by $\mathbf{T}=\{t_i\}_{i=1, ..., H\times W}$. Stage-I outputs a probability map $\mathbf{P}^{(1)}=f(\mathbf{I};\bm{\Theta}^{(1)})\subset{\mathbb{R}^{H\times W \times |\mathcal{L}|}}$ for each label at every pixel, where the probability density function $f(\cdot; \bm{\Theta}^{(1)})$ is a segmentation network parameterized by $\bm\Theta^{(1)}$. We use FCN \cite{Long2015} with reverse connections,  which is explained in Sec. \ref{sec:ReverseConnection}, as $\bm\Theta^{(1)}$. FCN is the backbone network throughout the paper. Each element $p^{(1)}_{i,l} \in \mathbf{P}^{(1)}$ is the probability that the $i$-th pixel in the input slice belongs to label $l$, where $l=0$ is the background, and $l=1, ..., |\mathcal{L}|$ are target organs. We define $p_{i,l}^{(1)}=\sigma(a_{i,l}^{(1)})=\frac{\exp(a_{i,l}^{(1)})}{\sum_{t=0}^{|\mathcal{L}|}\exp(a_{i,t}^{(1)})}$, where $a_{i,l}^{(1)}$ is the activation value of the $i$-th pixel on the $l$-th channel dimension. Let $\mathbf{A}^{(1)}=\{a_{i,l}^{(1)}\}_{i=1,...,H\times W, l=0,...,|\mathcal{L}|}$ be the activation map. The objective function to minimize for $\bm\Theta^{(1)}$ is given by
\begin{equation}
\label{eq:J1}
\mathcal{J}^{(1)}(\bm\Theta^{(1)})=-\frac{1}{H\times W}\left [ \sum_{i=1}^{H\times W}\sum_{j=0}^{|\mathcal{L}|}\mathbf{1}\left ( t_{i}=l \right )\log p^{(1)}_{i,l} \right ],
\end{equation}
where $\mathbf{1}(\cdot)$ is an indicator function.

Using a preliminary organ segmentation map to guide the computation of a better organ segmentation can be thought as employing an attentional mechanism. Towards this end, we propose an organ-attention module by
\begin{equation}
\label{eq:OAM}
\mathbf{Q} = \mathbf{W}* \mathbf{P}^{(1)} + \mathbf{b},
\end{equation}
where $*$ denotes the convolution operator, $\mathbf{W}$ indicates the convolutional filters, and $\mathbf{b}$ is the bias. (\ref{eq:OAM}) embeds cross-organ information into a single organ-attention map, $\mathbf{Q}$, which learns discriminative spatial attention for different organs automatically. By combining $\mathbf{Q}$ with the original input $\mathbf{I}$, we get an image which emphasizes each organ by
\begin{equation}
\mathbf{I}^{(2)} = \mathbf{I} \star {\mathbf{Q}},
\end{equation}
where $\star$ is the element-wise product operator. We apply $\mathbf{I}^{(2)}$ to the input of stage-II, and the probability of stage-II then becomes $\mathbf{P}^{(2)}=f(\mathbf{I}^{(2)};\bm\Theta^{(2)})$.

In order to drive stage-II to focus on organ regions without needing to deal with complicated non-local background, we define a selection function, $\mathbf{1}(\mathbf{P}^{(1)}_0  \leqslant \rho)$ where  $\mathbf{P}^{(1)}_0  =\{{p}_{i,0}^{(1)}\}_{i = 1, ..., H\times W}$ is the probability map provided by stage-I. In stage-II, we only accept the region if $p_{i,0}^{(1)}> \rho$ and do not back-propagate it to stage-I. The loss function for stage-II is formulated as
\begin{equation}
\mathcal{J}^{(2)}(\bm\Theta^{(2)},\mathbf{W}, \mathbf{b})=-\frac{1}{H\times W}\left [ \sum_{i=1}^{H\times W}\sum_{j=0}^{|\mathcal{L}|}\mathbf{1}\left(p_{i,0}^{(1)} \leqslant \rho\right )\cdot\mathbf{1}\left ( t_{i}=l \right )\log p^{(2)}_{i,l} \right ].
\end{equation}

To jointly optimize stage-I and stage-II, we define a loss function aiming at estimating parameters $\bm\Theta^{(1)}$, $\bm\Theta^{(2)}$, $\mathbf{W}$, and $\mathbf{b}$ by optimizing
\begin{equation}
\mathcal{J} = h^{(1)}\mathcal{J}^{(1)}(\bm\Theta^{(1)}) + h^{(2)}\mathcal{J}^{(2)}(\bm\Theta^{(2)},\mathbf{W}, \mathbf{b}),
\end{equation}
where $h^{(1)}$ and $h^{(2)}$ are the fusion weights.

\subsection{Reverse Connections}
\label{sec:ReverseConnection}

\begin{figure}[ht]
\centering
\includegraphics[width=0.9\textwidth]{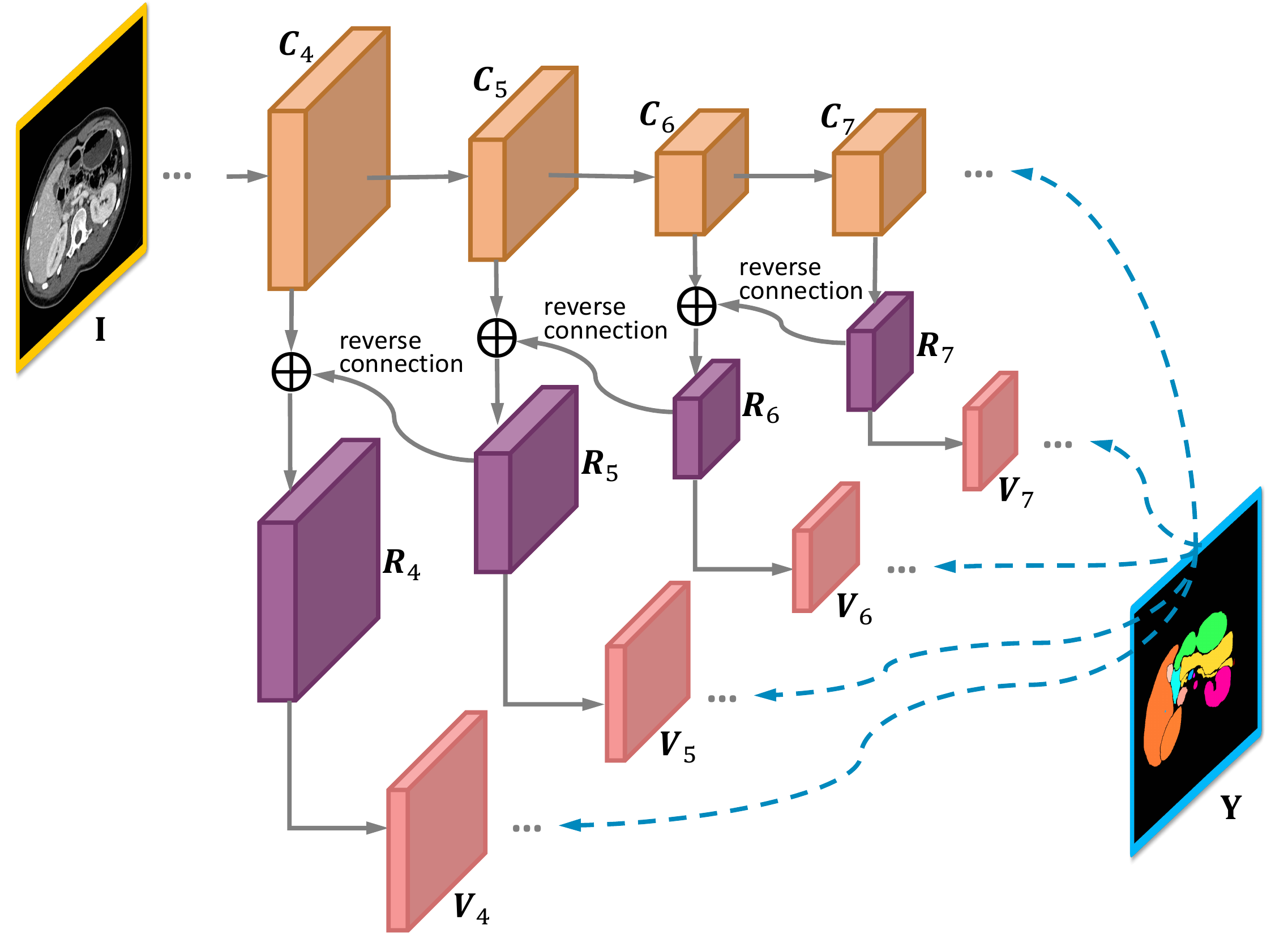}
\caption{The reverse connections architecture of OAN stage-I. The network has reverse connections to the output of convolutional layers. In the training step, both backbone network and reverse connection side-outputs are supervised by the ground-truth. Finally, all reverse connection side-outputs and the output of backbone network are fused and made to approach ground-truth.}
\label{fig:rf_overview}
\end{figure}

\begin{figure}[ht]
\centering
\includegraphics[width=0.6\textwidth]{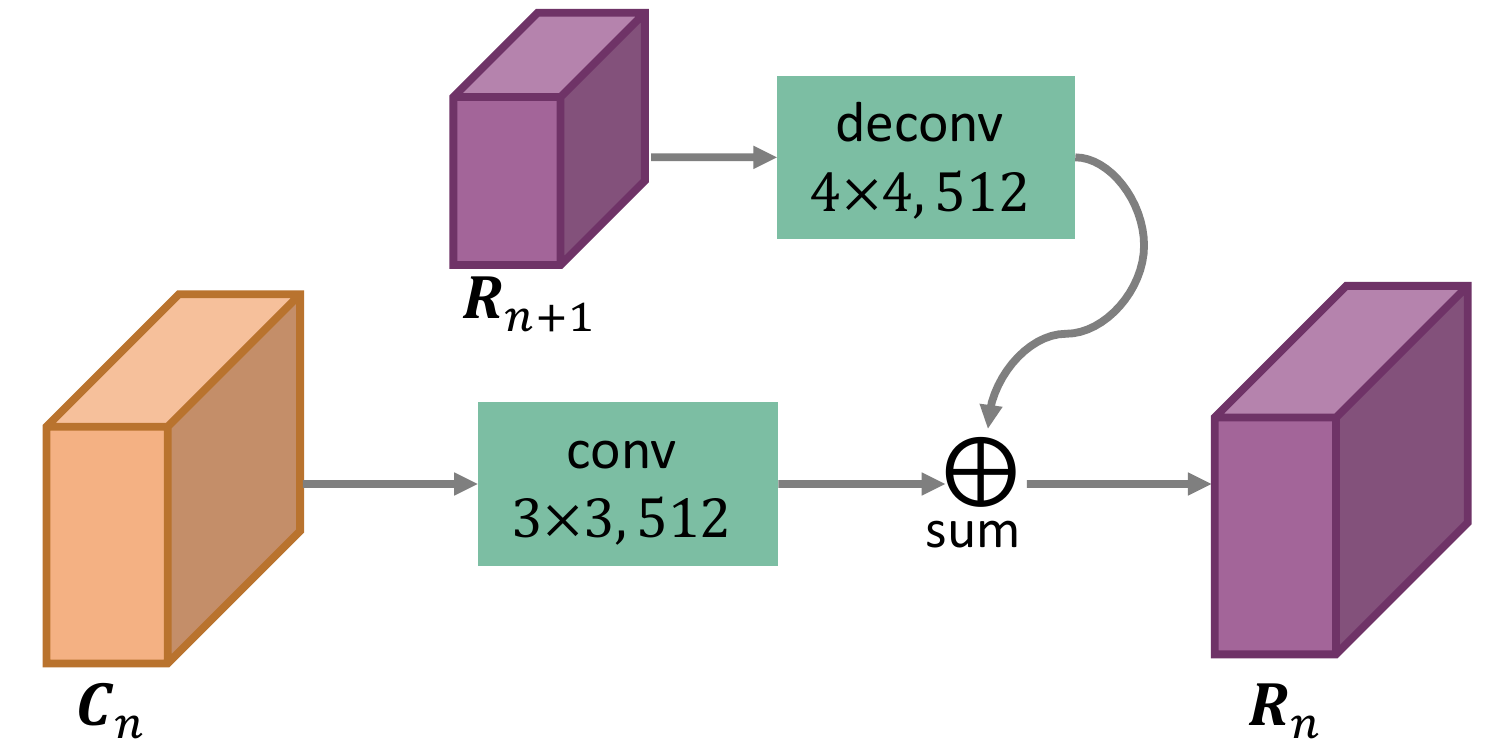}
\caption{A reverse connection block.}
\label{fig:rf_block}
\end{figure}

FCNs \cite{Long2015} have shown good segmentation results in recent studies, especially for single organ segmentation. However, for multi-organ segmentation, lower layers typically lack semantic information, which may lead to inaccurate segmentation particularly for smaller structures. Therefore, we propose reverse connections which feed coarse-scale (high) layer information backward to fine-scale (low) layer for semantic segmentation of multi-scale structures, inspired by \cite{Kong17}. This enables us to connect abstract high-level semantic information to the more detailed lower layers so that all the target organs have similar levels of details and abstract information at the same layer. The reverse connections framework for stage-I is shown in Fig.~\ref{fig:rf_overview}. Fig.~\ref{fig:rf_block} illustrates a reverse connection block. Let $\mathbf{R}_n$ denote the reverse connection map of the $n$-th convolutional layer in the backbone network, i.e. FCN in this study, where $\mathbf{C}_n$ is the output of the $n$-th convolutional layer. A convolutional layer (with $512$ channels by $3\times 3$ kernels) is added after $\mathbf{C}_n$, and a deconvolutional layer (with $512$ channels by $4\times 4$ kernels) is applied after $\mathbf{R}_{n+1}$. $\mathbf{R}_n$ is then obtained via an element-wise summation of these two maps. $\mathbf{R}_7$ is the output of a convolutional layer (with $512$ channels by $2\times 2$ kernels) grafted onto $\mathbf{C}_7$. Let $\mathbf{w}^n$ denote the corresponding weights for obtaining $\mathbf{R}_n$. Following \cite{Kong17}, we add reverse connections from $\mathbf{C}_4$ to $\mathbf{C}_7$.

With these learnable reverse connections, the semantic information of the lower layers can be enriched. In order to drive learned reverse connection maps to produce segmentation results approaching the ground-truth, we make each reverse connection map associate with a classifier. As the side-output layers proposed in \cite{Kong17} are designed for detection purposes, they are not suitable for our task. Instead we follow the side-outputs used in \cite{Xie2015}. More specifically, a convolutional layer (with $|\mathcal{L}|$ channels by $1\times 1$ kernels) is added on top of $\mathbf{R}_n$, whose output is denoted as $\mathbf{V}_n$, and followed by a deconvolutional layer (with $|\mathcal{L}|$ channels). We denote the weights of the $n$-th side-output layer by $\bm\theta^n$. The loss function for side-output layers $\mathcal{J}^{(s, 1)}$ is defined as
\begin{equation}
\mathcal{J}^{(s, 1)}(\bm\Theta^{(1)}, \mathbf{w}, \bm\theta) = \sum_{n=4}^7 h_{n}^{(s,1)}\ell_n^{(s,1)}\left(\bm\Theta^{(1)}, \mathbf{w}^n,\bm\theta^n\right),
\end{equation}
where $\ell_n^{(s,1)}=-\frac{1}{H\times W}\left [ \sum_{i=1}^{H\times W}\sum_{j=0}^{|\mathcal{L}|}\mathbf{1}\left ( t_{i}=l \right )\log p_{i,l}^{(s,1)} \right ]$ and $p^{(s,1)}_{i,l}$ is the probability output of the $n$-th side-output layer.

\begin{figure}[ht]
\centering
\includegraphics[width=0.8\textwidth]{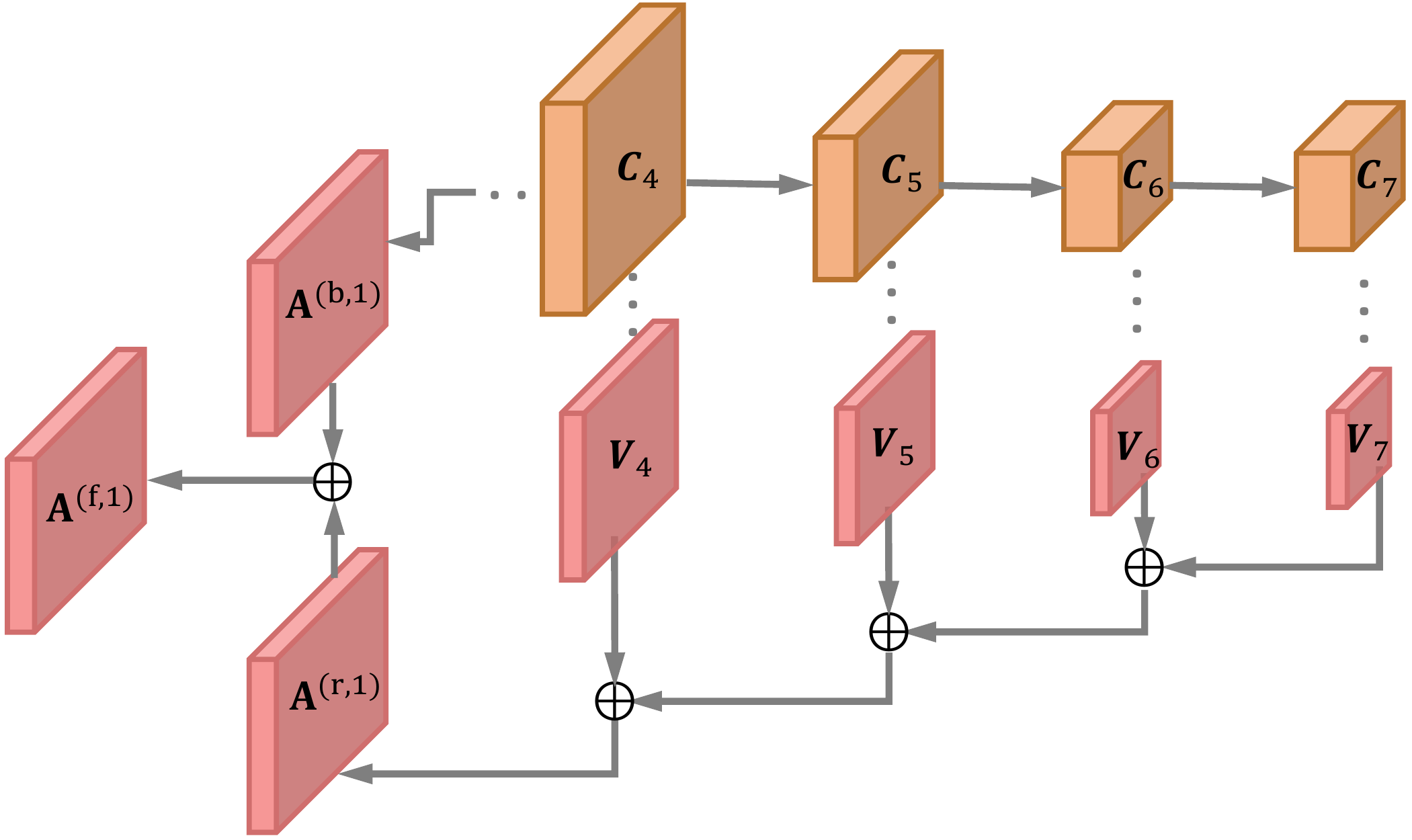}
\caption{Feature fusion strategy. A deep-to-shallow refinement is adopted for multi-scale side-output features. The final activation map ($\mathbf{A}^{(f,1)}$) for stage-I is an element-wise addition of the side-output activation map ($\mathbf{A}^{(r,1)}$) and the backbone network activation map ($\mathbf{A}^{(b,1)}$).}
\label{fig:merge}
\end{figure}

In order to combine the learned reverse connection maps of fine layers and coarse layers, we add up the predictions (i.e., $\mathbf{V}_n$) of the reverse connection maps from high layer to low layer gradually. First, $\mathbf{V}_6$ is fused with a $2\times$ upsampling of $\mathbf{V}_7$ by an element-wisely addition. Then we follow the same strategy and gradually merge $\mathbf{V}_5$ and $\mathbf{V}_4$, as shown in Fig.~\ref{fig:merge}. To obtain a fused activation map $\mathbf{A}^{(f,1)}=\{a_{i,l}^{(f,1)}\}_{i=1,...,H\times W, l=0, ..., |\mathcal{L}|}$ from the activation map of both side-outputs (i.e., $\mathbf{A}^{(r,1)}$) and convolutional layers in the backbone network (i.e., $\mathbf{A}^{(b,1)}$), a scale function is adopted followed by an element-wise addition by
\begin{equation}
\mathbf{A}^{(f,1)}_l = h^{(r,1)}_l \mathbf{A}^{(r,1)}_l + h^{(b,1)}_l\mathbf{A}^{(b,1)}_l, \qquad l = 0, ..., |\mathcal{L}|
\end{equation}
where $\mathbf{A}_l$ indicates the $l$-th channel of the activation map. $h^{(r,1)}_l $ and $h^{(b,1)}_l$ are fusion weights. Then the fused probability map, $\mathbf{P}^{(f,1)}=\{p_{i,l}^{(f,1)}\}_{i=1,...,H\times W, l=0,...,|\mathcal{L}|}$, can be obtained by $p^{(f,1)}_{i,l} = \sigma(a^{(f,1)}_{i,l})$.
The final objective function for stage-I is defined by
\begin{equation}
\begin{split}
&\mathcal{J}^{(1)}(\bm\Theta^{(1)}, \mathbf{w}, \bm\theta) = h^{(b,1)}\mathcal{J}^{(b,1)}(\bm\Theta^{(1)}) \\
&+ h^{(s,1)}\mathcal{J}^{(s,1)}(\bm\Theta^{(1)},\mathbf{w},\bm\theta) + h^{(f,1)}\mathcal{J}^{(f, 1)}(\bm\Theta^{(1)}, \mathbf{w}, \bm\theta),
\end{split}
\end{equation}
where $h^{(b,1)}$, $h^{(s,1)}$ and $h^{(f,1)}$ are fusion weights, and
\begin{equation}
\mathcal{J}^{(f, 1)}(\bm\Theta^{(1)}, \mathbf{w}, \bm\theta)=-\frac{1}{H\times W}\left [ \sum_{i=1}^{H\times W}\sum_{j=0}^{|\mathcal{L}|}\mathbf{1}\left ( t_{i}=l \right )\log p^{(f,1)}_{i,l} \right ].
\end{equation}

Note that in our full system with the two-stage organ-attention network and reverse connections, all the parameters are optimized simultaneously by standard back-propagation
\begin{equation}
\begin{split}
&(\hat{\bm\Theta}^{(1)},\hat{\mathbf{w}},\hat{\bm{\theta}},\hat{\bm\Theta}^{(2)},\hat{\mathbf{W}},\hat{\mathbf{b}}) \\
&= \arg\min\{\mathcal{J}^{(1)}(\bm\Theta^{(1)}, \mathbf{w}, \bm{\theta})+ h^{(2)}\mathcal{J}^{(2)}(\bm\Theta^{(2)}, \mathbf{W},\mathbf{b})\}.
\end{split}
\end{equation}

\subsection{Testing Phase}
\label{sec:Testing}

In the testing stage, given a slice $\mathbf{I}$, we obtain the stage-I and stage-II probability map by
\begin{equation}
\begin{split}
&\mathbf{P}^{(1)} = f(\mathbf{I}; \hat{\bm\Theta}^{(1)}, \hat{\mathbf{w}}, \hat{\bm\theta}) \\
&\mathbf{P}^{(2)} = f(\mathbf{I}; \hat{\bm\Theta}^{(2)}, \hat{\mathbf{W}},\hat{\mathbf{b}}),
\end{split}
\end{equation}
where $f(\cdot,\cdot)$ is the network functions defined in Sec. \ref{sec:TwoStageOAN}. A fused probability map of $\mathbf{P}^{(1)} $ and $\mathbf{P}^{(2)}$ is then given by
\begin{equation}
\mathbf{P} = \mathbf{P}^{(1)} \circ \mathbf{1}(\mathbf{P}_0^{(1)}>\rho) + \mathbf{P}^{(2)} \circ \mathbf{1}(\mathbf{P}_0^{(1)}\leqslant\rho).
\end{equation}
The final label map $\mathbf{S}=\{s_{i}\}_{i=1,...,H\times W}$ is determined by $s_i = \arg\min_{l\in\mathcal{L}} p_{i,l}$.

\section{Statistical Label Fusion Based on Local Structural Similarity}
\label{sec:Fusion}

\begin{figure*}
    \centering
    \includegraphics[width=1.0\linewidth]{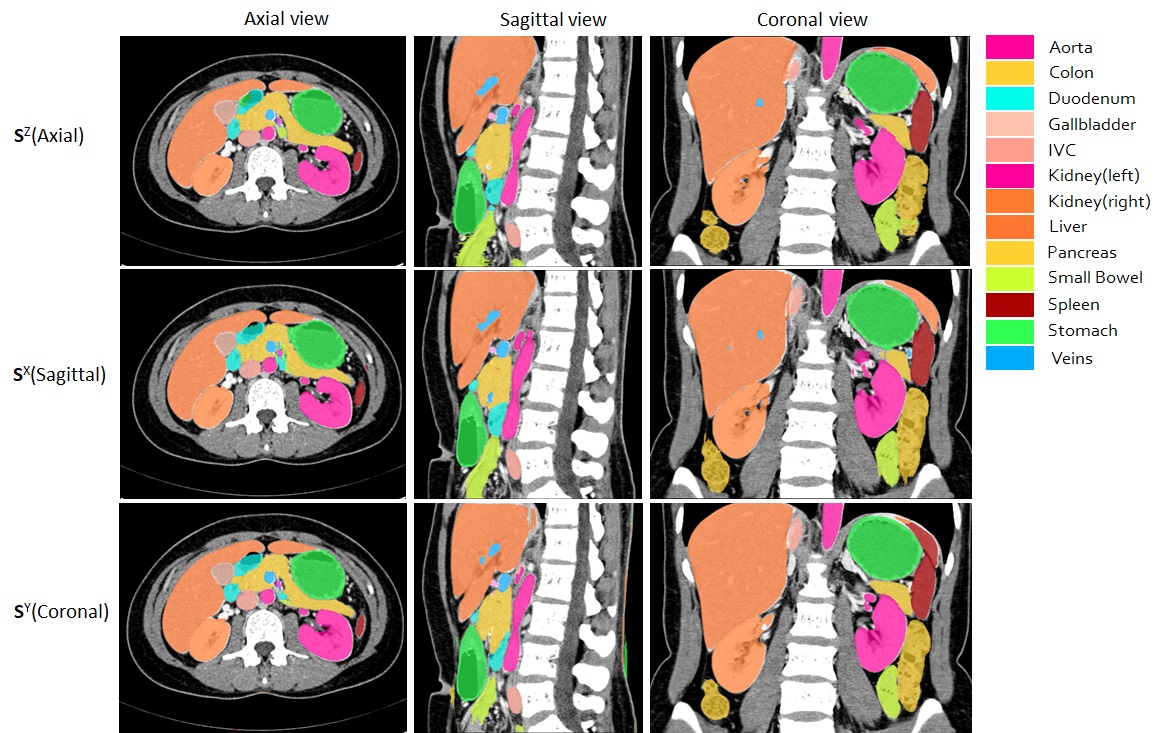}
    \caption{An example of multi-planar reconstruction view of OAN-RC estimations}
\label{fig:FCNResultMPR}
\end{figure*}

As described in Sec. \ref{sec:Introduction}, our OAN-RC is based on 2D images which is an extreme case of 3D anisotropic patches. In this section, we propose to fuse anisotropic information obtained from different viewing directions using isotropic 3D local patches to estimate the final segmentation. Let us denote the segmentation results by $\mathbf{S}^{j}, (j=1, \ldots, M=3)$, which are obtained as described in Sec. \ref{sec:Testing} from the axial (Z), sagittal (X), and coronal (Y) OAN-RCs. Depending on the viewing directions, sectional images contain different structures and may have different texture patterns in the same organs. These differences can cause nonidentical segmentations by the deep network as shown in Fig.~\ref{fig:FCNResultMPR} in 3D. In addition, there is no guarantee of connectivity between neighbor slices by independent use of slices for training and testing. Possible na\"{\i}ve approaches for determining the final segmentation in 3D from the OAN-RC results can be boolean operations such as union or intersection. Majority voting (MV) is another candidate for efficient fusion, however, theses approaches assume the same global weights of OAN-RC results. From the observations that the performance level of segmentation, e.g. sensitivity, can be different from viewing directions for each organ, we set the performance level to be an unknown variable when computing the probability of labeling. This concept is similar to the label fusion algorithms using expectation-maximization (EM) framework such as STAPLE (simultaneous truth and performance level estimation) and its extensions \cite{Warfield2004,Asman2012,Asman2013}.

Let us denote the true label of the $V$ by $\mathbf{T}$, which is unknown, and the unknown performance level parameter of segmentation by $\theta$. The segmentations from the deep networks $\mathbf{S} = \left\{\mathbf{S}^{j}|j=1,...,M\right\}$ are observed values.
Under this condition, the basic EM framework is performed by following two steps in an iterative manner: 1) to compute $Q^0(\theta|\theta^{(k)}) = E_{\mathbf{T}}\left[\ln L(\theta|\mathbf{S}, \mathbf{T}) | \mathbf{S},\theta^{(k)}\right]$ which is the expected value of the log likelihood, $\ln L(\theta | \mathbf{S}, \mathbf{T}) = \ln P(\mathbf{S}, \mathbf{T}|\theta)$, under the current estimate of the parameters  $\theta^{(k)}$ at $k^{th}$ iteration, and
2) to find the parameter $\theta^{(k+1)}$ which maximizes $Q^0(\theta|\theta^{(k)})$.

The maximization step can be written as
\begin{equation} \label{eq:Q}
\begin{split}
&\theta^{(k+1)} = \arg \max_{\theta} E_{\mathbf{T}} \left[\ln P(\mathbf{S}, \mathbf{T} |\theta) | \mathbf{S},\theta^{(k)}\right] \\
& = \arg \max_{\theta} E_{\mathbf{T}} \left[\ln P(\mathbf{S} | \mathbf{T}, \theta)P(\mathbf{T}) | \mathbf{S},\theta^{(k)}\right] \\
& = \arg \max_{\theta} \sum_{\mathbf{T}} \ln \left\{P(\mathbf{S} | \mathbf{T}, \theta)P(\mathbf{T})\right\} P(\mathbf{T} | \mathbf{S},\theta^{(k)}) \\
& = \arg \max_{\theta} \sum_{\mathbf{T}} \left\{\ln P(\mathbf{S} | \mathbf{T}, \theta) + \ln P(\mathbf{T}) \right\}P(\mathbf{T}|\mathbf{S},\theta^{(k)}).
\end{split}
\end{equation}
By assuming independence between $\mathbf{T}$ and $\theta$ in our problem, the second term $\sum_{\mathbf{T}} \ln P(\mathbf{T})P(\mathbf{T}|\mathbf{S},\theta^{(k)})$ in (\ref{eq:Q}) becomes free of $\theta$ and the maximization step can be written as
\begin{equation} \label{eq:Qprime}
\begin{split}
\theta^{(k+1)} & = \arg \max_{\theta} \sum_{\mathbf{T}} \ln P(\mathbf{S}| \mathbf{T}, \theta)P(\mathbf{T}|\mathbf{S},\theta^{(k)}) \\
               & = \arg \max_{\theta} E_{\mathbf{T}} \left[ \ln P(\mathbf{S} | \mathbf{T}, \theta) | \mathbf{S}, \theta^{(k)} \right].
\end{split}
\end{equation}
Therefore, we redefine $Q^{0}(\theta|\theta^{(k)})$ as $Q(\theta|\theta^{(k)}) = E_{\mathbf{T}} \left[ \ln P(\mathbf{S} | \mathbf{T}, \theta) | \mathbf{S}, \theta^{(k)} \right]$.

\begin{figure}
  \centering
  \includegraphics[width=1.0\linewidth]{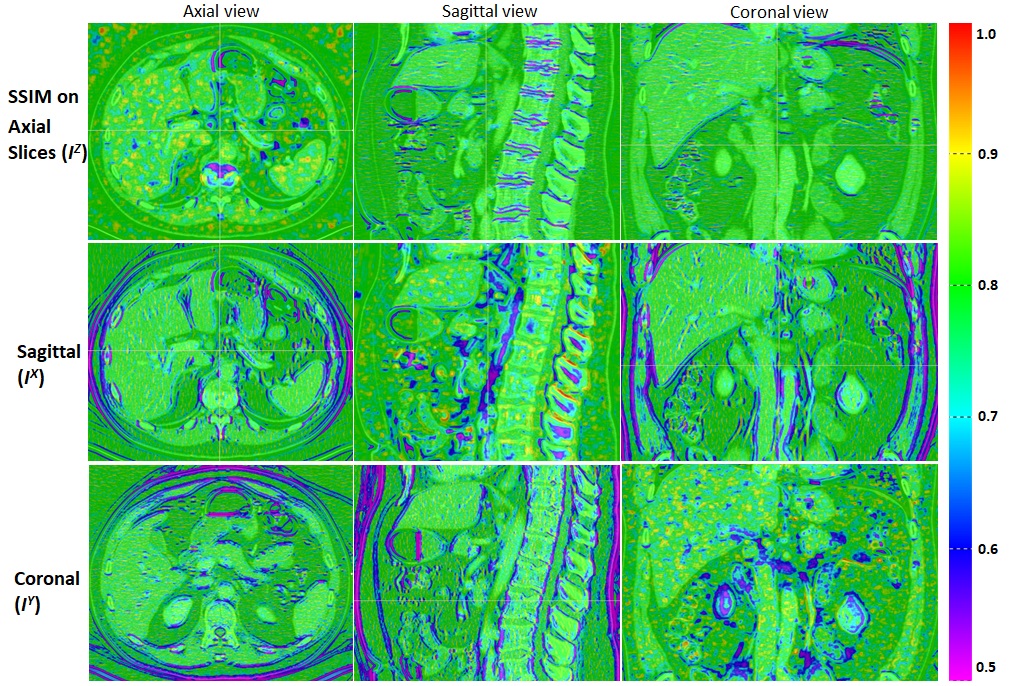}
  \caption{The local structural similarity map between 2D slices and the 3D volume. Each row is captured from the same similarity map computed on one viewing direction. Each column shows the captures images at the same location computed from different viewing directions.}
  \label{fig:SSIMMap}
\end{figure}

The performance level parameter in this framework is a global property representing the overall confidence of deep network segmentation for the whole volume. However, it can also vary according to the voxel spatial locations via the local and neighbor structures as we use 2D slices for the initial segmentation. Therefore, we propose to combine local structural similarity shown from a specific viewing direction to the original 3D volume and the global performance level, conceptually similar to local weighted voting \cite{Sabuncu2010}. We compute the probability of correspondence between 2D images and the 3D volume by structural similarity (SSIM) \cite{Wang2004} by
\begin{equation} \label{eq:Correspondence}
    \begin{split}
    \alpha_{i}^{j} & = P\left({ \ell_{2}(I_{i}^{j}) | \ell_{3}(V_{i}) }\right) \equiv SSIM\left(\ell_{2}(I_{i}^{j}), \ell_{3}(V_{i})\right) \\
                         &= \frac{\left(2 \mu_{\ell_{2}(I_{i}^{j})} \mu_{\ell_{3}(V_{i})} + c_{1} \right) \left(2 \sigma_{\ell_{2}(I_{i}^{j}) \ell_{3}(V_{i})} + c_{2}\right)}  {\left(\mu_{\ell_{2}(I_{i}^{j})}^{2} + \mu_{\ell_{3}(V_{i})}^{2} + c_{1}\right) \left(\sigma_{\ell_{2}(I_{i}^{j})}^{2} + \sigma_{\ell_{3}(V_{i})}^{2} + c_{2}\right)},
    \end{split}
\end{equation}
where $\alpha_{i}^{j}$ is the SSIM from the $j^{th}$ viewing direction at the $i^{th}$ voxel. $c_{1}$ and $c_{2}$ are user-defined constants, and $\ell_{2}(I_{i})$ and $\ell_{3}(V_{i})$ represent local 2D and 3D patches centered at the $i^{th}$ voxel, respectively. $\mu_{\ell}$ and $\sigma_{\ell}$ are the average and standard deviation of the patch $\ell$, respectively, and $\sigma_{\ell_{2}(I_{i}) \ell_{3}(V_{i})}$ is the covariance of $\ell_{2}(I_{i})$ and $\ell_{3}(V_{i})$. Fig. \ref{fig:SSIMMap} shows an example of the structural similarity computed on different viewing directions as a color map.

Considering the local image properties, the expectation of log likelihood function in our problem becomes
\begin{equation} \label{eq:LogLikelihood}
    \begin{split}
    Q\left( \mathbf{\theta} | \mathbf{\theta}^{(k)} \right) & = E \left[ \ln P\left( \mathbf{S},I|\mathbf{T},V,\mathbf{\theta} \right) | \mathbf{S}, I, V, \mathbf{\theta}^{(k)} \right] \\
    & = \sum_{\mathbf{T}} \ln P\left(\mathbf{S},I|\mathbf{T},V,\mathbf{\theta}\right) P\left( \mathbf{T} | \mathbf{S}, I, V, \mathbf{\theta}^{(k)} \right).
    \end{split}
\end{equation}
The global underlying performance level parameters of the deep network segmentations is defined as
\begin{equation} \label{eq:PerformanceLevelDef}
    \theta_{js's} \equiv P \left( \mathbf{S}_{i}^{j}=s'| \mathbf{T}_{i}=s, \theta_{js's}^{(k)} \right),
\end{equation}
where $\theta_{js's}$ is the probability of the voxel labeled as $s'$ from the $j^{th}$ deep network with the current estimated performance value $\theta_{js's}^{(k)}$, when the true label is $s$.

To make the problem simple, we assume conditional independence between labeling and the original volume intensities.
The labeling probability with the target image intensity then becomes
\begin{equation} \label{eq:LabelProbability}
    \begin{split}
    &P\left({\mathbf{S}_{i}^{j}=s', \ell_{2}(I_{i}^{j}) | \mathbf{T}_{i}=s,\ell_{3}(V_{i}),\theta_{js's}^{(k)} }\right) \\
    & = P\left({\mathbf{S}_{i}^{j}=s' | \mathbf{T}_{i}=s, \theta_{js's}^{(k)}} \right) P\left({\ell_{2}(I_{i}^{j}) | \ell_{3}(V_{i})}\right)\\
    & = \theta_{js's} \alpha_{i}^{j}.
    \end{split}
\end{equation}

\subsection{E-step}

In the expectation step (E-step), we estimate the probability of voxelwise labels.
Let us denote the probability that the true label of $i^{th}$ voxel is $s \in \mathcal{L}$ at the $k^{th}$ iteration by $\omega_{si}^{(k)}$. When the deep network segmentations $\mathbf{S}$ and performance level parameters at the $k^{th}$ iteration $\mathbf{\theta}^{(k)}$ are given, $\omega_{si}^{(k)}$ can be then described as
\begin{equation} \label{eq:Likelihood}
    P\left(\mathbf{T}_{i}=s | \mathbf{S}, I, V, \mathbf{\theta}^{(k)}\right) \equiv \omega_{si}^{(k)},
\end{equation}
where $\mathbf{\theta} \in \mathbb{R}^{N \times |\mathcal{L}| \times |\mathcal{L}|}$ is the vector of all $(\theta_{js's})^{T}$.
From the independence between $\mathbf{S}^{X},~\mathbf{S}^{Y}$, and $\mathbf{S}^{Z}$, we apply Bayesian theorem to (\ref{eq:Likelihood}).
\begin{equation} \label{eq:Weight}
    {\omega_{si}^{(k)}} = \frac{ P({\mathbf{T}_{i} = s}) \prod_{j} P \left({\mathbf{S}_{i}^{j} = s', \ell_{2}(I_{i}^{j})| \mathbf{T}_{i} = s, \ell_{3}(V_{i}), \mathbf{\theta}_{j}^{(k)} }\right)}
                   { \sum_{n} {P(\mathbf{T}_{i} = n) \prod_{j} P\left( \mathbf{S}_{i}^{j} = s', \ell_{2}(I_{i}^{j}) | \mathbf{T}_{i} = n, \ell_{3}(V_{i}), \mathbf{\theta}_{j}^{(k)} \right)} },
\end{equation}
where $P(T_{i} = s)$ is a \textit{priori} of the $i^{\mathrm{th}}$ voxel.
By applying (\ref{eq:LabelProbability}) to (\ref{eq:Weight}), we then obtain the probability of voxelwise labeling as
\begin{equation} \label{eq:WeightApp}
    {\omega_{si}^{(k)}} = \frac {P(\mathbf{T}_{i} = s) \prod_{j} {\theta_{js's}^{(k)} \alpha_{i}^{j}}}
                         {\sum_{n} P(\mathbf{T}_{i}=n) \prod_{j} {\theta_{js'n}^{(k)} \alpha_{i}^{j}}}.
\end{equation}

\subsection{M-step}

In the maximization step (\textit{M-step}), the goal is to find the performance parameters, $\mathbf{\theta}$, which maximize (\ref{eq:LogLikelihood}) with the current given parameters.
Considering each $\mathbf{S}^{j}$ and $\theta_{j}$ independently, the expectation of log likelihood function in (\ref{eq:LogLikelihood}) can be expressed with the estimated voxelwise probability in \textit{E-step}. Then the performance parameter of each segmentation can be formulated to find the solution which maximizes the summation of voxelwise probability as
\begin{equation}
\label{eq:MaxQ}
     \theta_{j}^{(k+1)} = \arg \max_{\theta_{j}} Q\left(\theta_{j} | \theta_{j}^{(k)} \right) = \arg \max_{\theta_{j}} \sum_{i} Q_{i} \left(\theta_{j} | \theta_{j}^{(k)} \right),
\end{equation}
where $Q_{i} = {E [ \ln P( \mathbf{S}_{i},\ell_{2}(I_{i})|\mathbf{T}_{i},\ell_{3}(V_{i}),\theta^{(k)} ) | \mathbf{S},I, V, \theta^{(k)} ]}$ at $i^{th}$ voxel. By applying (\ref{eq:Likelihood}) and (\ref{eq:LabelProbability}), (\ref{eq:MaxQ}) becomes
\begin{equation}
\label{eq:DerivationTheta}
    \begin{split}
     \theta_{j}^{(k+1)} &= \arg \max_{\theta_{j}} \sum_{i}\sum_{s} P(\mathbf{T}_{i}=s|\mathbf{S},I,V,{\theta}^{(k)}) \times \ln {P\left(\mathbf{S}_{i}^{j}, \ell_{2}(I_{i}^{j}) | \mathbf{T}_{i}=s, \ell_{3}(V_{i}), {\theta}_{j}^{(k)} \right)} \\
     &= \arg \max_{\theta_{j}} \sum_{i}\sum_{s} \omega_{si}^{(k)} \ln {P\left(\mathbf{S}_{i}^{j}, \ell_{2}(I_{i}^{j}) | \mathbf{T}_{i}=s, \ell_{3}(V_{i}), {\theta}_{j}^{(k)} \right)} \\
     &= \arg \max_{\theta_{j}} \sum_{s'}\sum_{i:\mathbf{S}_{i}^{j}=s'}\sum_{s} \omega_{si}^{(k)} \times ln P\left( \mathbf{S}_{i}^{j}=s', \ell_{2}(I_{i}^{j}) | \mathbf{T}_{i}=s, \ell_{3}(V_{i}), {\theta}_{j}^{(k)} \right) \\
     &= \arg \max_{\theta_{j}} \sum_{s'}\sum_{i:\mathbf{S}_{i}^{j}=s'}\sum_{s} \omega_{si}^{(k)} \ln \theta_{js's}\alpha_{i}^{j}.
    \end{split}
\end{equation}

From the definition of $\theta$ in (\ref{eq:PerformanceLevelDef}), the summation of probability mass function, $\sum_{s'}\theta_{js's}^{(k)}$,  must be $1$, and (\ref{eq:MaxQ}) becomes a constrained optimization problem which can be solved by introducing a Lagrange multiplier, $\lambda$. We then obtain the optimal solution by making the first gradient zero as

\begin{equation}\label{eq:derivative}
    0 = {{\partial} \over{\partial\theta_{js's}}} \left[ {Q\left(\theta_{j} | \theta_{j}^{(k)}\right) + \lambda \sum_{s'} \theta_{js's}} \right].
\end{equation}
By applying the derivation of $Q$ in (\ref{eq:LogLikelihood}), (\ref{eq:MaxQ}) and (\ref{eq:DerivationTheta}), (\ref{eq:derivative}) becomes

\begin{equation}\label{eq:derivation}
\begin{split}
  &0 = {{\sum_{i:\mathbf{S}_{i}^{j}=s'} \omega_{si}^{(k)} \alpha_{i}^{j} } \over {\theta_{js's}}} + \lambda \\
  & {\theta}_{js's}^{(k+1)} = {{\sum_{i:\mathbf{S}_{i}^{j}=s'} \omega_{si}^{(k)} \alpha_{i}^{j} } \over {- \lambda}}.
\end{split}
\end{equation}
By substituting the constraint of $\sum_{s'}\theta_{js's}^{(k)} = 1$, we can obtain the final optimal solution as
\begin{equation}
\label{eq:PerformanceLevelResult}
    {\theta}_{js's}^{(k+1)} = {{\sum_{i:\mathbf{S}_{i}^{j}=s'} \alpha_{i}^{j} \omega_{si}^{(k)}} \over {\sum_{i}\omega_{si}^{(k)}}}.
\end{equation}

The two steps, (\ref{eq:WeightApp}) and (\ref{eq:PerformanceLevelResult}), are then computed alternatively in the EM iterations until they converge. From the final values of (\ref{eq:WeightApp}), the final segmentation can be computed by graph-based approaches such as \cite{Boykov2001}.

\subsection{Parallel computing using GPUs}

The fusion step can be efficiently computed in a  parallel way on a GPU. The local structural similarity $\alpha_{i}^{j}$ of $i$-th voxel in $j$th deep network and \textit{priori} $P(T_{i})$ can be computed for each voxel and saved as a pre-processing step. In the EM iterations, as shown in (\ref{eq:WeightApp}), the probability can be computed and updated for each structure at each voxel. In our implementation, a GPU thread is logically allocated for each voxel. However, to reduce the used memory and computation cost, the target volume of interest (VOI) for each structure $s$ is computed in an extended region as $\delta=4$ voxels for each direction from $V(\bigcup_{j}\mathbf{S}^{j}=s)$ in our implementation. For parallel computing, one CPU thread is allocated to a structure and launches a kernel of one GPU to compute EM iteration for each structure.

\section{Experimental Results}
\label{sec:Experiments}

We evaluated our methods on $236$ abdominal CT images of normal cases under an IRB (Institutional Review Board) approved protocol in Johns Hopkins Hospital as a part of the FELIX project for pancreatic cancer research \cite{Lugo-Fagundo17}. CT images were obtained by Siemens Healthineers (Erlangen,Germany) SOMATOM Sensation and Definition CT scanners. CT scans are composed of $(319-1051)$ slices of $(512 \times 512)$ images, and have voxel spatial resolution of $\left([0.523-0.977] \times [0.523-0.977] \times 0.5 \right)mm^{3}$. All CT scans are contrast enhanced images and obtained in the portal venous phase.

A total of $13$ structures for each case were segmented by four human annotators/raters, one case by one person, and confirmed by an independent senior expert. The structures include the aorta, colon, duodenum, gallbladder, interior vena cava (IVC), kidney (left, right), liver, pancreas, small bowel, spleen, stomach, and large veins. Vascular structures were segmented only outside of the organs in order to make the structures exclusive to each other (i.e. no overlaps).

As explained in Sec. \ref{sec:OAN-RC}, we used OAN-RCs for multi-organ segmentation whose backbone FCNs had been pre-trained by $PascalVOC$ dataset \cite{Everingham12}. From the possible variants of FCNs (e.g., FCN-32s, FCN-16s, and FCN-8s), which depend on how they combine the fine detailed predictions \cite{Shelhamer17}, we selected FCN-8s in this study because it captures very fine details in the $3^{rd}$ and $4^{th}$ pooling layer, and keeps high-level semantic contextual information from the final layer. Our algorithm was implemented and tested on a workstation with Intel i7-6850K CPU, NVidia TITAN X (PASCAL) GPU. With $236$ cases, the initial segmentations using OAN-RCs were tested by four-fold cross-validation. All the input images of OAN-RCs are $1.5$ times enlarged by upsampling, which lead to improved performance in our experiments.

In the fusion step, the average probability of $\mathbf{S}^{X}, \mathbf{S}^{Y}, \mathbf{S}^{Z}$ are taken as a \textit{priors} in (\ref{eq:WeightApp}) and the initial performance levels $\theta_{js's}^{(0)}$ were computed by randomly selecting 5 cases and by comparing them to the ground-truth. To compute the local patch-based structural similarity in (\ref{eq:Correspondence}), patches of $(4.5 \times 4.5 \times 4.5) mm^3$ size cubes were used for 3D volume. Since CT voxels are not always isotropic and spatial resolutions can be different between scan volumes, we re-sampled the 3D patch with $0.5mm$ length cubic voxels so that the same size of $(9 \times 9 \times 9)$ 3D patches and $(9 \times 9)$ 2D patches from all directions can be used for all cases in our experiments.

\begin{figure}
  \centering
  \includegraphics[width=1.0\linewidth]{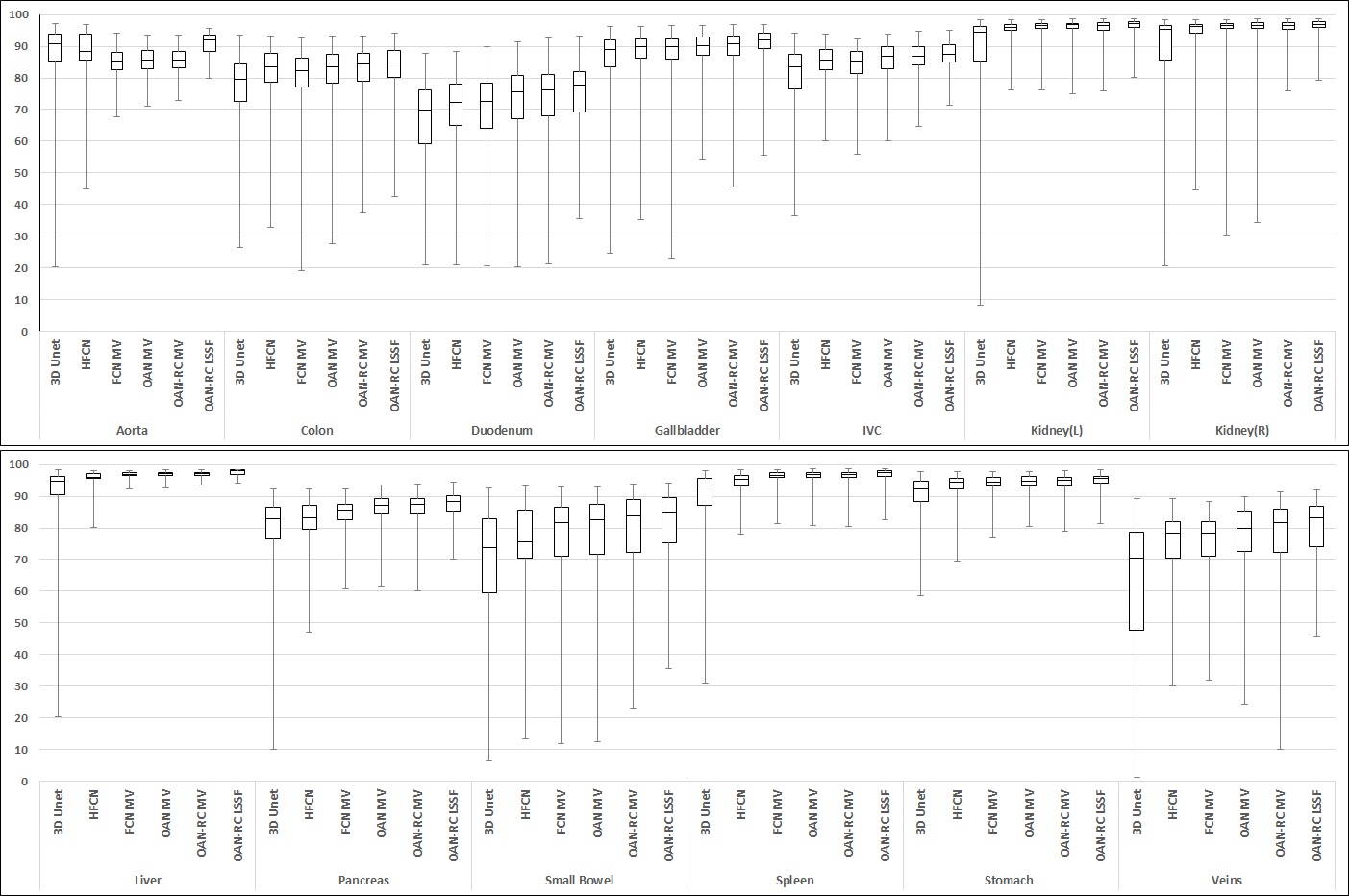}
  \caption{Box plots of the Dice-S{\o}rensen similarity coefficients of 13 structures to compare performance. As in typical box plots, the box represents the first quartile, median, and the third quartile from the lower border, middle and the upper boarder, respectively, and the lower and the upper whiskers show the minimum and the maximum values. (LSSF: Local Similarity-based Statistical Fusion.) }
  \label{fig:EvaluationAllOrgans}
\end{figure}

The final segmentation results using OAN-RC with local structural similarity-based statistical fusion (LSSF) were compared with the 3D-patch based state-of-the-art approaches, 3D Unet\cite{Cicek2016} and hierarchical 3D FCN (HFCN) \cite{Roth2017} as well as 2D-based FCN, OAN and OAN-RC with majority voting (MV). For a quantitative comparison, we computed the well-known Dice-S{\o}rensen similarity coefficient (DSC) and the  surface distances based on the manual annotations as ground-truth. For a structure $s$, DSC is computed as ${{2V(\mathbf{S}=s \bigcap \mathbf{T}=s)} \over {V(\mathbf{S}=s) + V(\mathbf{T}=s)}}$ where $\mathbf{S}$ is the estimated segmentation and $\mathbf{T}$ is the ground-truth, i.e. manual annotations in this study. The surface distance was computed from each vertex of the ground-truth and to the estimates of our algorithms. Fig. \ref{fig:EvaluationAllOrgans} shows comparison results by box plots, while Tables \ref{tab:EvaluationDICE} and \ref{tab:EvaluationDistance} represent the mean and standard deviations for all the $236$ cases. 

As shown in Fig. \ref{fig:EvaluationAllOrgans}, the basic OAN-RC outperforms other state-of-the-art approaches and our local structural similarity-based fusion improves the results even more. We note that although DSC shows the relative overall volume similarity, it does not quantify the boundary smoothness or the boundary noise of the results. But evaluating the surface distances, see below, shows that our method works effectively for both the whole volumes and the boundaries of the organs.

\begin{table*}[t]
\caption{DICE-S{\o}rensen similarity coefficient (DSC, \%) of thirteen segmented organs. (mean $\pm$ standard deviation of 236 cases)}
\label{tab:EvaluationDICE}
\centering
\footnotesize
\begin{tabular}{|c|c|c|c|c|c|c|}
\hline
Structure&3D U-net&HFCN&FCN MV&OAN MV&OAN-RC MV&OAN-RC LSSF\\
\hline
Aorta& 87.0$\pm$12.3& 88.3$\pm$ 8.8& 85.0$\pm$4.2 & 85.5$\pm$ 4.2 & 85.3$\pm$ 4.1&\textbf{91.8}$\pm$ 3.5\\
\hline
Colon& 77.0$\pm$11.0& 79.3$\pm$ 9.2& 80.3$\pm$ 9.1 & 81.5$\pm$ 9.4 & 82.0$\pm$ 8.8&\textbf{83.0}$\pm$ 7.4\\
\hline
Duodenum& 66.8$\pm$12.8 & 70.3$\pm$ 10.4 & 70.2$\pm$11.3& 72.6$\pm$11.4& 73.4$\pm$11.1&\textbf{75.4}$\pm$ 9.1\\
\hline
Gallbladder& 85.4$\pm$10.3 & 87.9$\pm$ 7.5& 87.8$\pm$ 8.3& 88.9$\pm$ 6.2& 89.4$\pm$ 6.1&\textbf{90.5}$\pm$ 5.3\\
\hline
IVC& 80.8$\pm$10.2 & 84.7$\pm$ 5.9& 84.0$\pm$ 6.0& 85.6$\pm$ 5.8& 86.0$\pm$ 5.5&\textbf{87.0}$\pm$ 4.2\\
\hline
Kidney(L)& 83.9$\pm$22.4 & 95.2$\pm$ 2.6& 96.1$\pm$ 2.0& 96.2$\pm$ 2.2& 95.9$\pm$ 2.3&\textbf{96.8}$\pm$ 1.9\\
\hline
Kidney(R)& 88.0$\pm$14.4 & 95.6$\pm$ 4.5& 95.8$\pm$ 4.9& 95.9$\pm$ 4.9& 96.0$\pm$ 2.5&\textbf{98.4}$\pm$ 2.1\\
\hline
Liver& 91.4$\pm$ 9.9 & 95.7$\pm$ 1.8& 96.8$\pm$ 0.8& 97.0$\pm$ 0.9& 97.0$\pm$ 0.8&\textbf{98.0}$\pm$ 0.7\\
\hline
Pancreas& 79.3$\pm$11.7 & 81.4$\pm$10.8& 84.3$\pm$ 4.9 & 86.2$\pm$ 4.5& 86.6$\pm$ 4.3&\textbf{87.8}$\pm$ 3.1\\
\hline
Small bowel& 69.9$\pm$17.3 & 71.1$\pm$15.0& 76.9$\pm$14.0& 78.0$\pm$13.8& 79.0$\pm$13.4&\textbf{80.1}$\pm$10.2\\
\hline
Spleen& 89.6$\pm$9.5 & 93.1$\pm$ 2.1& 96.3$\pm$ 1.9& 96.4$\pm$ 1.9& 96.4$\pm$ 1.7&\textbf{97.1}$\pm$ 1.5\\
\hline
Stomach& 90.1$\pm$ 7.2& 93.2$\pm$ 5.4 & 93.9$\pm$ 3.2& 94.2$\pm$ 2.9& 94.2$\pm$ 3.0&\textbf{95.2}$\pm$ 2.6\\
\hline
Veins& 60.7$\pm$23.7& 74.5$\pm$10.5& 74.8$\pm$10.7& 76.8$\pm$11.2& 77.4$\pm$12.1&\textbf{80.7}$\pm$ 9.3\\
\hline
\end{tabular}
\end{table*}

\begin{table*}[t]
\caption{Average surface distances of thirteen segmented organs for all 236 cases. (mean $\pm$ standard deviation of average surface distances in $mm$)}
\label{tab:EvaluationDistance}
\centering
\footnotesize
\begin{tabular}{|c|c|c|c|c|c|c|}
\hline
Structure & 3D U-net & HFCN & FCN MV & OAN MV & OAN-RC MV & OAN-RC LSSF \\
\hline
Aorta & 0.44 $\pm$1.01& 0.42$\pm$0.58& 0.56$\pm$0.47& 0.47$\pm$0.42& 0.44$\pm$0.28 & \textbf{0.39}$\pm$0.21\\
\hline
Colon & 6.75$\pm$9.01& 6,35$\pm$8.12& 6.27$\pm$7.44& 5.65$\pm$7.25& 4.07$\pm$5.72& \textbf{3.59}$\pm$4.17\\
\hline
Duodenum & 2.01$\pm$2.46& 1.70$\pm$2.18& 1.71$\pm$2.25& 1.49$\pm$1.87& 1.54$\pm$1.43& \textbf{1.36}$\pm$1.31\\
\hline
Gallbladder & 1.31$\pm$0.76& 1.21$\pm$0.50& 1.22$\pm$0.52& 1.12$\pm$0.50& 1.05$\pm$0.41& \textbf{0.95}$\pm$0.37\\
\hline
IVC & 1.57$\pm$1.53& 1.15$\pm$1.05& 1.26$\pm$1.08& 1.16$\pm$1.38& 1.12$\pm$1.24& \textbf{1.08}$\pm$1.03\\
\hline
Kidney(L) & 0.77$\pm$1.04& 0.41$\pm$0.42& 0.36$\pm$0.47& 0.34$\pm$0.47& 0.30$\pm$0.33& \textbf{0.30}$\pm$0.30\\
\hline
Kidney(R) & 1.39$\pm$2.01& 1.03$\pm$1.68& 1.05$\pm$1.74& 0.74$\pm$1.32& 0.54$\pm$1.09& \textbf{0.45}$\pm$0.89\\
\hline
Liver & 1.89$\pm$3.21& 1.60$\pm$ 0& 1.61$\pm$2.98& 1.39$\pm$2.64& 1.32$\pm$1.74& \textbf{1.23}$\pm$1.52\\
\hline
Pancreas & 1.78$\pm$1.05& 1.51$\pm$0.80& 1.41$\pm$0.88& 1.19$\pm$0.82& 1.17$\pm$0.72& \textbf{1.05}$\pm$0.65\\
\hline
Small bowel & 4.21$\pm$5.78& 4.01$\pm$6.01& 3.91$\pm$6.05& 3.20$\pm$4.05& 3.37$\pm$5.48& \textbf{3.01}$\pm$3.35\\
\hline
Spleen & 0.98$\pm$0.56& 0.59$\pm$0.37& 0.60$\pm$0.36& 0.56$\pm$0.40& 0.47 $\pm$0.27& \textbf{0.42}$\pm$0.25\\
\hline
Stomach & 2.78$\pm$5.89 & 2.50$\pm$5.02& 2.51$\pm$5.13& 2.36$\pm$5.65& 1.88$\pm$1.64& \textbf{1.68}$\pm$1.55\\
\hline
Veins & 2.31$\pm$4.51& 1.75$\pm$3.51& 1.69$\pm$3.61& 1.92$\pm$6.48& 1.40$\pm$3.61& \textbf{1.21}$\pm$3.05\\
\hline
\end{tabular}
\end{table*}

Tables \ref{tab:EvaluationDICE} and \ref{tab:EvaluationDistance} represent the mean and standard deviations of performance measures for 13 critical organs. Similar to the box plots, they show that  our OAN-RCs with statistical fusion improves the overall mean performance and also reduces the standard deviations significantly.

The OAN-RC training and testing can be computed in parallel for each view direction. In our experiments, the training took $40$ hours for $120,000$ iterations for $177$ training cases and the average testing time for each volume was $76.73$ seconds. The fusion time depended on the volume of the target structure, and the average computation time for $13$ organs was 6.87 seconds.

\section{Discussion}


Multi-organ segmentation using OAN-RCs alone, without the statistical fusion, gave similar or better performance compared with the state-of-the-art approaches summarized in \cite{Karasawa2017}. In the specific case of the pancreas, state-of-the-art methods showed (mean $\pm$ standard deviations) segmentation accuracies as $74.4 \pm 20.2(\%)$ on 140 cases \cite{Saito2016}, $78.5 \pm 14.0 (\%)$ on 150 cases \cite{Karasawa2017}, $78.0 \pm 8.2 (\%)$ on $82$ cases \cite{Roth2016} and $75.74 \pm 10.47 (\%) $ (on the whole slice) versus $82.4 \pm 5.7 (\%)$ (reduced region of interest) on $82$ cases \cite{Zhou2017} in terms of DSC. We cannot make a direct comparison because in these datasets CT images and manual segmentations (i.e. annotation) for the ground-truth are different from each other. But our OAN-RCs segmentations on our larger dataset shows similar or better performances in terms of DSC. Among target organs, our performance on structures such as gallbladder and pancreas, whose sizes are relatively small and have particularly weak boundaries improves significantly from using basic FCNs or using OANs without reverse connections.

\begin{figure}
\centering
  \includegraphics[width=1.0\linewidth]{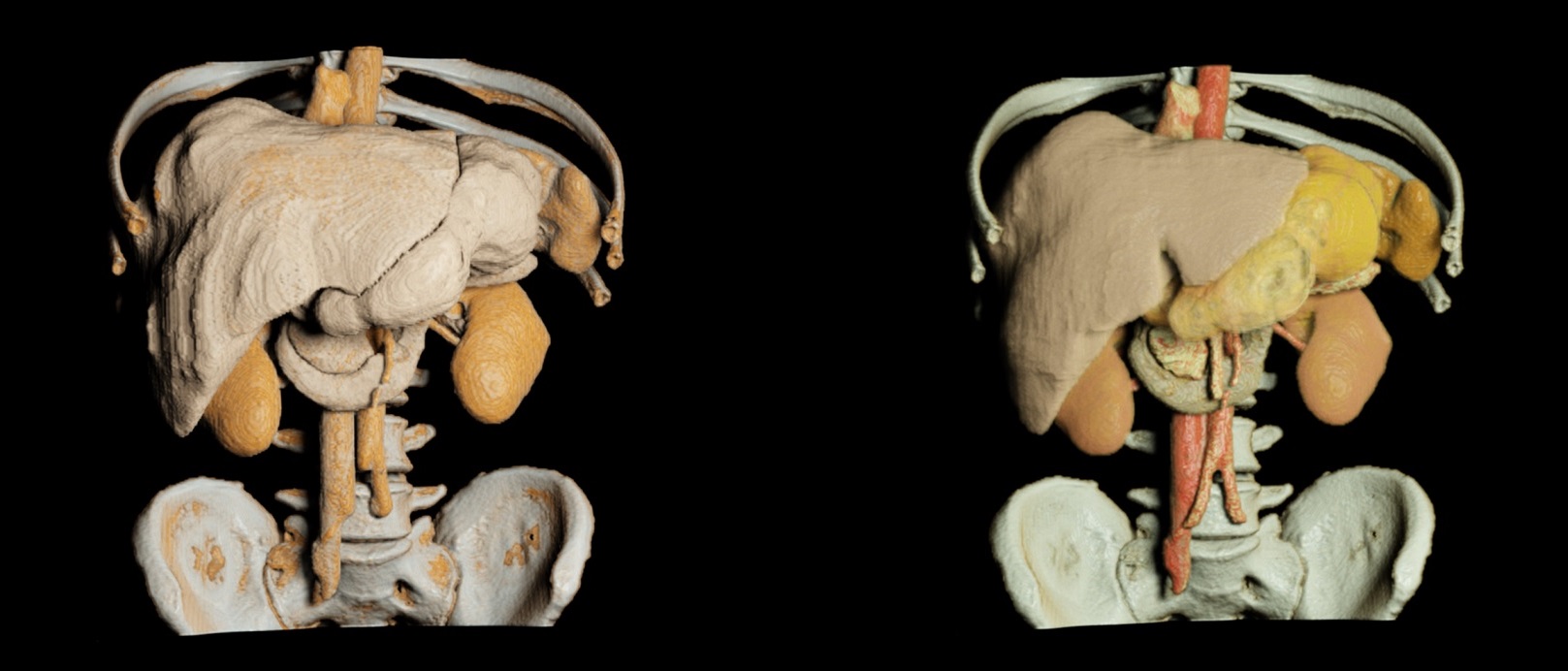}
  \caption{3D photo-realistic rendering of the ground-truth (left) and the results from OAN-RC with statistical fusion (right). The aorta, duodenum, IVC, liver, kidneys, pancreas, duodenum, spleen, and stomach are rendered. The difference between our results and the ground-truth are almost visually indistinguishable. To differentiate adjacent organs and from manual segmentation, different color setting were applied to the our methods results.}
  \label{fig:ManualvsOAN}
\end{figure}

\begin{figure*}
  \centering
  \footnotesize
  \includegraphics[width=1.0\linewidth]{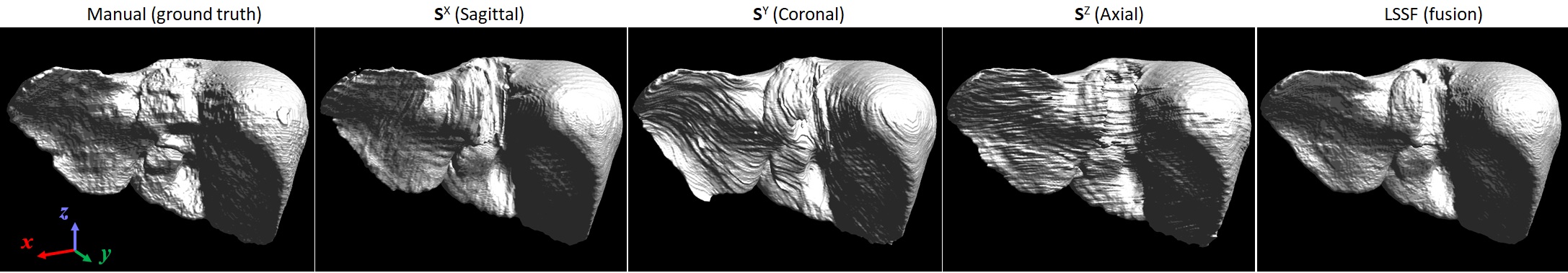}
  \includegraphics[width=1.0\linewidth]{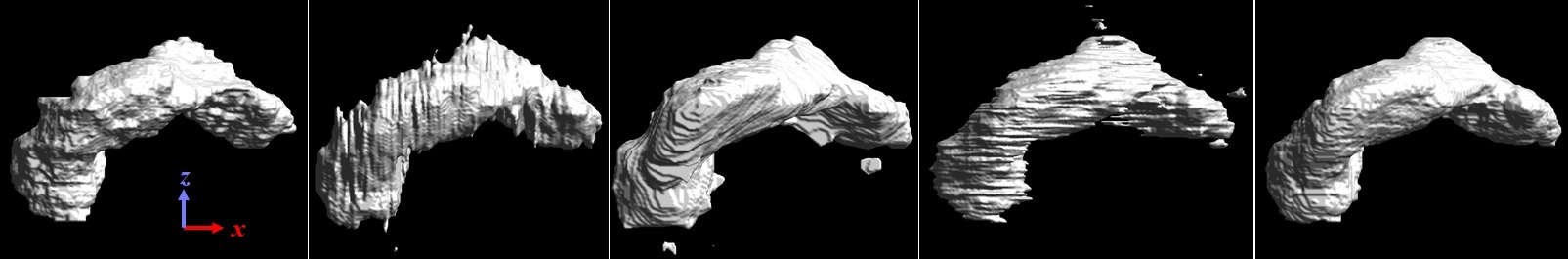}
  \parbox[t]{1.0\columnwidth}{\relax}
  \centering{\makebox[30pt]{(a) Liver (upper) and Pancreas (lower)}}
  \includegraphics[width=1.0\linewidth]{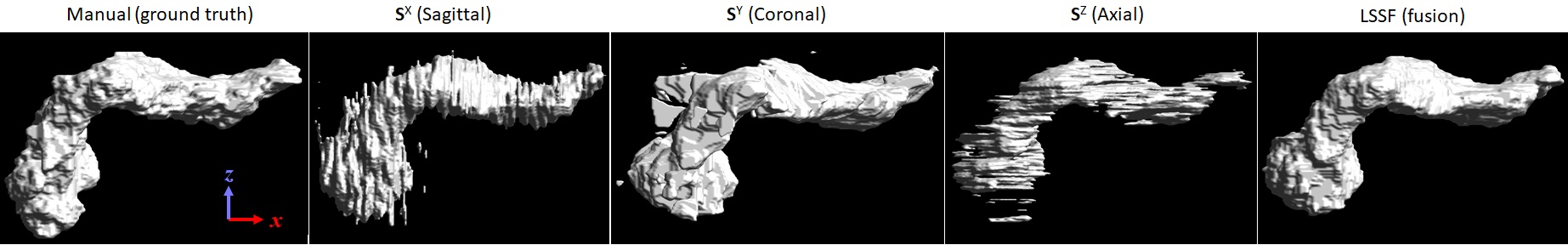}
  \includegraphics[width=1.0\linewidth]{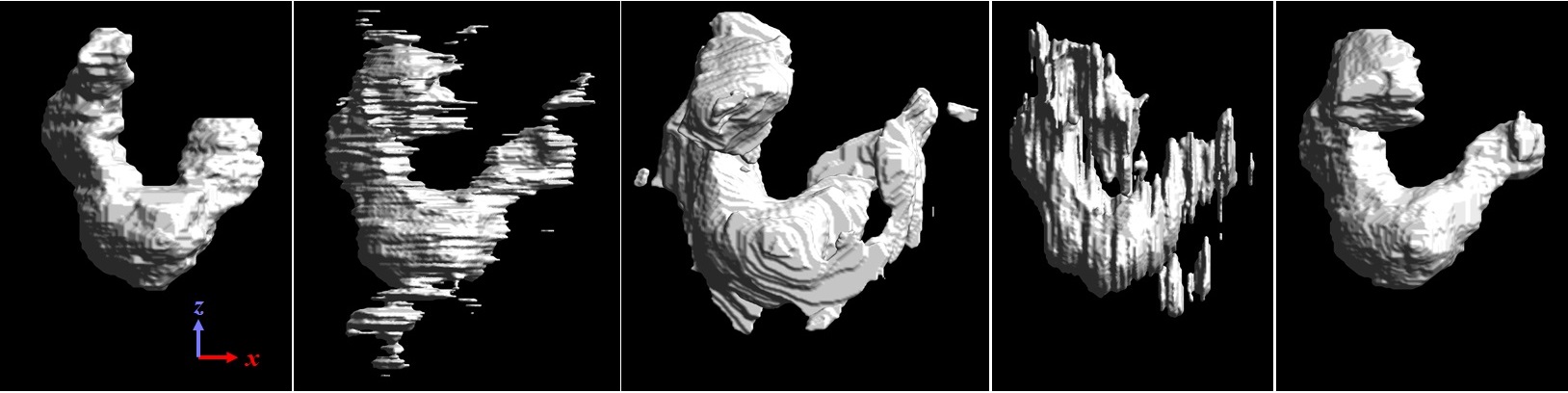}
  \parbox[t]{1.0\columnwidth}{\relax}
  \centering{\makebox[30pt]{(b) Pancreas (Upper) and Duodenum (lower)}}
   \caption{Effects of local structural similarity-based statistical fusion (LSSF) for estimating 3D surfaces. From left to right, the manual segmentation (ground-truth), initial segmentations from OAN-RCs with X, Y, Z slices, and the results of our proposed algorithm with statistical fusion. (a) When $\mathbf{S}^X$, $\mathbf{S}^Y$, and $\mathbf{S}^Z$ show similar result, statistical fusion produces smoother and less-noisy boundaries. (b) Surface estimation examples when initial OAN-RCs give differing results. But our approach effectively fuses the information, exploiting  the local structural similarity.}
  \label{fig:SurfaceGeneration}
\end{figure*}

\begin{figure*}
  \centering
  \footnotesize
  \includegraphics[width=1.0\linewidth]{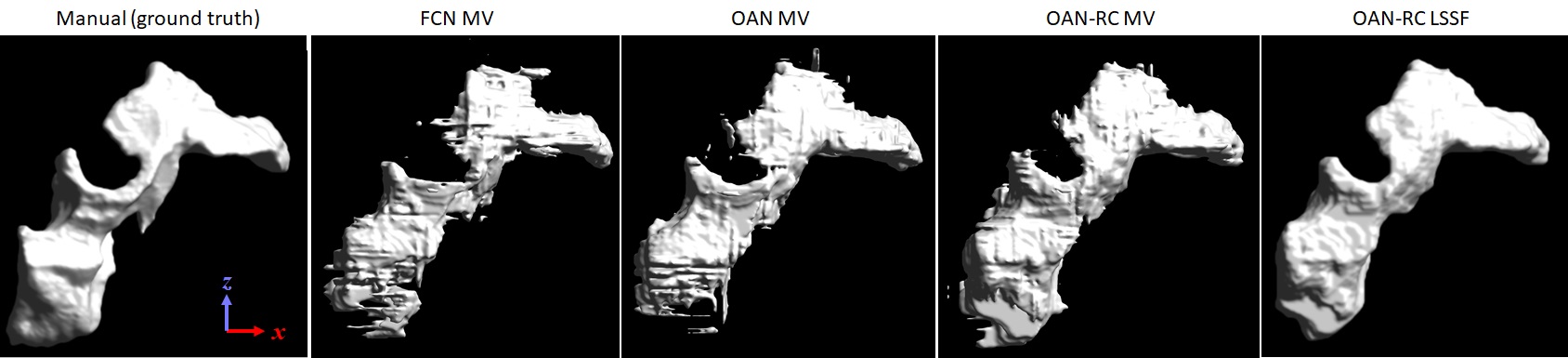}
  \parbox[t]{1.0\columnwidth}{\relax}
  \centering{\makebox[30pt]{(a)}}

  \includegraphics[width=1.0\linewidth]{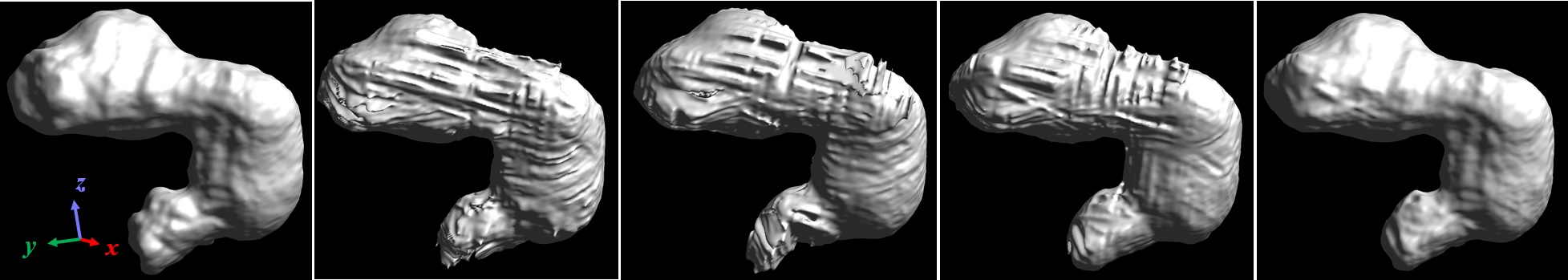}
  \parbox[t]{1.0\columnwidth}{\relax}
  \centering{\makebox[30pt]{(b)}}
  \caption{Examples of FCN, OAN, OAN-RC, and OAN-RC. The manual segmentation (ground-truth), FCN MV, OAN MV, OAN-RC MV, OAN-RC LSFF (from left to right). (a) Pancreas: DSC($\%$) and surface distances (mean$\pm$ standard deviation in $mm$) to the ground-truth are $72.5$ and $2.13\pm1.74$ (FCN MV), $77.2$ and $1.90\pm1.77$ (OAN MV), $82.4$ and $1.33\pm1.31$ (OAN-RC MV), and $85.5$ and $0.71\pm0.81$ (OAN-RC LSSF), respectively. (b) Stomach: DSC($\%$) and surface distances (mean$\pm$ standard deviation in $mm$) to the ground-truth are $92.5$ and $2.44\pm1.27$ (FCN MV), $93.6$ and $1.63\pm1.14$ (OAN MV), $94.9$ and $2.25\pm1.30$ (OAN-RC MV), and $97.1$ and $1.26\pm0.88$ (OAN-RC LSSF), respectively.}
  \label{fig:SurfaceGeneration2}
\end{figure*}

Moreover, as shown in Sec. \ref{sec:Experiments}, our statistical fusion based on local structural similarity improves the overall segmentation accuracies in terms of both DSC and average surface distances. In particular, there are significant performance improvements for the minimum values as shown in Fig. \ref{fig:EvaluationAllOrgans}, which helps explain the robustness of the algorithm. The differences can be depicted more clearly by visualizing the 3D surfaces as shown in Figs \ref{fig:SurfaceGeneration} - \ref{fig:SurfaceGeneration2}. The noise of the deep network segmentations is distributed over large regions, without much connectivity, and occasionally they show significantly different patterns. But our fusion step exploits structural similarity which outputs clean and smooth boundaries by effectively combining different information based on the local structure of the original 3D volume.

When applying our proposed method and interpreting the evaluation results, we must address several considerations.

As shown in our experiments, our proposed algorithm also outperforms 3D patch based approaches. But 3D (isotropic) patch-based approaches have several issues which make it hard to apply to this problem. To make bigger patch size, they require more parameters and hence require more training data or, if this is not available, significant data augmentation (.e.g, by scaling, rotation, and elastic deformation). In addition, there can be practical memory limitation on GPUs which restricts the expandable patch size. The limited patch size means that the deep networks receptive field sizes contains only limited local information which is problematic for multi-organ segmentation and the discontinuities between the patches also raises problems. It is possible that solutions to these three problems may make 3D patch based methods work better in the future. Unlike 3D approaches, the local structure-similarity used in our fusion method effectively combine the information from anisotropic patches to 3D at each voxel. Fig. \ref{fig:ManualvsOAN} shows an example generated by our proposed algorithm, which is visually indistinguishable from manual segmentation for almost all target structures. 

The ground-truth used in this study for training and evaluation was specified using manual annotations by human observers. It is well known that there can be significant inter-/intra-observer variations in manual segmentation. But, as explained before, the ground-truth was created by four human observers and checked by experts in a visual way, and we randomly divided testing groups in our 4-fold cross-validation to avoid biased comparison. However, it is still possible that inaccuracies due to human variability may affect the evaluation as well as the training. This can be further intensively explored as separate experiments.

Another possible consideration when applying the proposed approach is the image quality which can affect both of manual annotations and deep network segmentation results. Various factors such as spatial resolution, level of artifacts and reconstruction kernels should be considered. The dataset used in this study has been collected between $2005$ to $2009$ in the same institute with control over the scanning parameters. As explained in Sec. \ref{sec:Experiments}, the CT protocol is the portal venous phase and the spatial resolution is almost isotropic. But different scanning parameters and artifacts may affect our algorithms performance when applied to other datasets.

The same issues about manual segmentations and image qualities can be raised in general segmentation and evaluations. Specifically for our proposed approach, especially in the fusion step, the way of computing \textit{priori}, $P(T)$, used in (\ref{eq:WeightApp}) can in practice affect the final segmentation. But considering that the deep network segmentation results from different viewing-directions are independently obtained, the mean can be accepted in general. However, if the deep network segmentations show clear tendencies towards over-estimation or under-estimation, then different types of models for \textit{priors} may need to be used in order to improve the final result for practical applications.

One of the main advantages of our algorithm is the efficient computation time. The segmentation of $13$ organs of the whole volume takes similar to or less than 1 minute with better performance reported than the state-of-the-art methods \cite{Karasawa2017}. Hence our approach can be practically useful in clinical environments.

\section{Conclusion}
\label{sec:Conclusion}

In this paper, we proposed a novel framework for multi-organ segmentation using OAN-RCs with statistical fusion exploiting structural similarity. Our two-stage organ-attention network reduces uncertainties at weak boundaries, focuses attention on organ regions with simple context, and adjusts FCN error by training the combination of original images and OAMs. Reverse connections deliver abstract level semantic information to lower layers so that hidden layers can be assisted to contain more semantic information and give good results even for small organs. The results are improved by the statistical fusion, based on local structural similarity, which smooths our noise and removes biases leading to better overall segmentation performance in terms of DSC and surface distances. We showed that our performance is better than previous state of the art algorithms.
Our framework is not specific to any particular body region, but gives high quality and robust results for abdominal CTs, which are typically challenging regions due to their low contrast, large intra-/inter-variations, and  different scales. In addition, the efficient computational time of our algorithm makes our approach  practical for clinical environments such as CAD, CAS or RT.

\bibliography{library}

\begin{thebibliography}{10}

\bibitem{Asman2012}
A.~J. Asman and B.~A. Landman.
\newblock {Formulating spatially varying performance in the statistical fusion
  framework}.
\newblock {\em IEEE Transactions on Medical Imaging}, 31(6):1326--1336, 2012.

\bibitem{Asman2013}
A.~J. Asman and B.~A. Landman.
\newblock {Non-local statistical label fusion for multi-atlas segmentation}.
\newblock {\em Medical Image Analysis}, 17(2):194--208, 2013.

\bibitem{Boykov2001}
Y.~Boykov, O.~Veksler, and R.~Zabih.
\newblock {Fast approximate energy minimization via graph cuts}.
\newblock {\em IEEE Transactions on Pattern Analysis and Machine Intelligence},
  23(11):1222--1239, 2001.

\bibitem{Chen2017}
H.~Chen, Q.~Dou, L.~Yu, J.~Qin, and P.-A. Heng.
\newblock {VoxResNet : Deep voxelwise residual networks for brain segmentation
  from 3D MR images}.
\newblock {\em NeuroImage}, 2017.

\bibitem{Chen2016}
L.-C. Chen, G.~Papandreou, I.~Kokkinos, K.~Murphy, and A.~L. Yuille.
\newblock {DeepLab: Semantic Image Segmentation with Deep Convolutional Nets,
  Atrous Convolution, and Fully Connected CRFs}.
\newblock 2016.

\bibitem{chen2016attention}
L.-C. Chen, Y.~Yang, J.~Wang, W.~Xu, and A.~L. Yuille.
\newblock Attention to scale: Scale-aware semantic image segmentation.
\newblock In {\em Proceedings of the IEEE conference on computer vision and
  pattern recognition}, pages 3640--3649, 2016.

\bibitem{Chu2013}
C.~Chu, M.~Oda, T.~Kitasaka, K.~Misawa, M.~Fujiwara, Y.~Hayashi, Y.~Nimura,
  D.~Rueckert, and K.~Mori.
\newblock {Multi-organ segmentation based on spatially-divided probabilistic
  atlas from 3D abdominal CT images}.
\newblock {\em Lecture Notes in Computer Science}, 8150 LNCS(PART 2):165--172,
  2013.

\bibitem{Cicek2016}
{\"{O}}.~{\c{C}}i{\c{c}}ek, A.~Abdulkadir, S.~S. Lienkamp, T.~Brox, and
  O.~Ronneberger.
\newblock {3D U-net: Learning dense volumetric segmentation from sparse
  annotation}.
\newblock {\em Lecture Notes in Computer Science}, 9901 LNCS:424--432, 2016.

\bibitem{Dou2016}
Q.~Dou, H.~Chen, Y.~Jin, L.~Yu, J.~Qin, and P.~A. Heng.
\newblock {3D deeply supervised network for automatic liver segmentation from
  CT volumes}.
\newblock {\em Lecture Notes in Computer Science}, 9901 LNCS:149--157, 2016.

\bibitem{Everingham12}
M.~Everingham, L.~Van~Gool, C.~K.~I. Williams, J.~Winn, and A.~Zisserman.
\newblock "the {PASCAL} {V}isual {O}bject {C}lasses {C}hallenge 2012
  {(VOC2012)} {R}esults".
\newblock
  "http://www.pascal-network.org/challenges/VOC/voc2012/workshop/index.html".

\bibitem{Havaei2017}
M.~Havaei, A.~Davy, D.~Warde-Farley, A.~Biard, A.~Courville, Y.~Bengio, C.~Pal,
  P.-M. Jodoin, and H.~Larochelle.
\newblock {Brain Tumor Segmentation with Deep Neural Networks}.
\newblock {\em Medical Image Analysis}, 35:18--31, 2017.

\bibitem{Heimann2009}
T.~Heimann, B.~van Ginneken, M.~Styner, Y.~Arzhaeva, V.~Aurich, C.~Bauer,
  A.~Beck, C.~Becker, R.~Beichel, G.~Bekes, F.~Bello, G.~Binnig, H.~Bischof,
  A.~Bornik, P.~Cashman, Y.~Chi, A.~Cordova, B.~Dawant, M.~Fidrich, J.~Furst,
  D.~Furukawa, L.~Grenacher, J.~Hornegger, D.~Kainmüller, R.~Kitney,
  H.~Kobatake, H.~Lamecker, T.~Lange, J.~Lee, B.~Lennon, R.~Li, S.~Li,
  H.~Meinzer, G.~Nemeth, D.~Raicu, A.~Rau, E.~van Rikxoort, M.~Rousson,
  L.~Rusko, K.~Saddi, G.~Schmidt, D.~Seghers, A.~Shimizu, P.~Slagmolen,
  E.~Sorantin, G.~Soza, R.~Susomboon, J.~Waite, A.~Wimmer, and I.~Wolf.
\newblock {Comparison and evaluation of methods for liver segmentation from CT
  datasets}.
\newblock {\em IEEE Transactions on Medical Imaging}, 28:1251--1265, 2009.

\bibitem{Iglesias2015}
J.~E. Iglesias and M.~R. Sabuncu.
\newblock {Multi-atlas segmentation of biomedical images: A survey}.
\newblock {\em Medical Image Analysis}, 24(1):205--219, 2015.

\bibitem{Kamnitsas2017}
K.~Jamnitsas, C.~Ledig, V.~F. Newcombe, J.~P. Simpson, A.~D. Kane, D.~K. Menon,
  D.~Rueckert, and B.~Glocker.
\newblock {Efficient multi-scale 3D CNN with fully connected CRF for accurate
  brain lesion segmentation}.
\newblock {\em Medical Image Analysis}, 36:61--78, 2017.

\bibitem{Kada2015}
T.~Kada, M.~G. Linguraru, M.~Hori, R.~M. Summers, N.~Tomiyama, and Y.~Sato.
\newblock {Abdominal multi-organ segmentation from CT images using conditional
  shape-location and unsupervised intensity priors}.
\newblock {\em Medical Image Analysis}, 26(1):1--18, 2015.

\bibitem{Karasawa2017}
K.~Karasawa, M.~Oda, T.~Kitasaka, K.~Misawa, M.~Fujiwara, C.~Chu, G.~Zheng,
  D.~Rueckert, and K.~Mori.
\newblock {Multi-atlas pancreas segmentation: Atlas selection based on vessel
  structure}.
\newblock {\em Medical Image Analysis}, 39:18--28, 2017.

\bibitem{Kirbas2004}
C.~Kirbas and F.~Quek.
\newblock A review of vessel extraction techniques and algorithms.
\newblock {\em ACM Computing Surveys}, 36:81--121, 2004.

\bibitem{Kong17}
T.~Kong, F.~Sun, A.~Yao, H.~Liu, M.~Lu, and Y.~Chen.
\newblock {RON:} reverse connection with objectness prior networks for object
  detection.
\newblock In {\em {IEEE} Conference on Computer Vision and Pattern Recognition
  (CVPR)}, 2017.

\bibitem{Lesage2009}
D.~Lesage, E.~D. Angelini, I.~Bloch, and G.~Funka-Lea.
\newblock A review of 3d vessel lumen segmentation techniques: Models, features
  and extraction schemes.
\newblock {\em MEdical Image Analysis}, 13:819--845, 2009.

\bibitem{Li2015}
G.~Li, X.~Chen, F.~Shi, W.~Zhu, J.~Tian, and D.~Xiang.
\newblock Automatic liver segmentation based on shape constraints and
  deformable graph cut in ct images.
\newblock {\em IEEE Transactions on Image Processing}, 24:5315--5329, 2015.

\bibitem{Long2015}
J.~Long, E.~Shelhamer, and T.~Darrell.
\newblock Fully convolutional networks for semantic segmentation.
\newblock {\em Proceedings of the IEEE Conference on Computer Vision and
  Pattern Recognition}, pages 3431--3440, 2015.

\bibitem{Lugo-Fagundo17}
C.~Lugo-Fagundo, B.~Vogelstein, A.~Yuille, and E.~K. Fishman.
\newblock Deep learning in radiology: Now the real work begins.
\newblock {\em Journal of the American College of Radiology}, 15:364--367,
  2018.

\bibitem{Mharib2012}
A.~M. Mharib, A.~R. Ramli, S.~Mashohor, and R.~B. Mahmood.
\newblock Survey on liver ct image segmentation methods.
\newblock {\em Artificial Intelligence Review}, 37, 2012.

\bibitem{Milletari2016}
F.~Milletari, N.~Navab, and S.~A. Ahmadi.
\newblock {V-Net: Fully convolutional neural networks for volumetric medical
  image segmentation}.
\newblock {\em Proceedings - 2016 4th International Conference on 3D Vision,
  3DV 2016}, pages 565--571, 2016.

\bibitem{Nascimento2016}
J.~Nascimento and G.~Carneiro.
\newblock {Multi-atlas segmentation using manifold learning with deep belief
  networks}.
\newblock {\em Proceedings - International Symposium on Biomedical Imaging},
  2016-June, 2016.

\bibitem{Rajchl2017a}
M.~Rajchl, M.~C. Lee, O.~Oktay, K.~Kamnitsas, J.~Passerat-Palmbach, W.~Bai,
  M.~Damodaram, M.~A. Rutherford, J.~V. Hajnal, B.~Kainz, and D.~Rueckert.
\newblock {DeepCut: Object Segmentation from Bounding Box Annotations Using
  Convolutional Neural Networks}.
\newblock {\em IEEE Transactions on Medical Imaging}, 36(2):674--683, 2017.

\bibitem{Roth2018}
H.~Roth, M.~Oda, N.~Shimizu, H.~Oda, Y.~Hayashi, T.~Kitasaka, M.~Fujiwara,
  K.~Misawa, and K.~Mori.
\newblock Towards dense volumetric pancreas segmentation in ct using 3d fully
  convolutional networks.
\newblock 2017.

\bibitem{Roth2015}
H.~R. Roth, A.~Farag, L.~Lu, E.~B. Turkbey, and R.~M. Summers.
\newblock {Deep convolutional networks for pancreas segmentation in CT
  imaging}.
\newblock 2015.

\bibitem{Roth2016}
H.~R. Roth, L.~Lu, A.~Farag, A.~Sohn, and R.~M. Summers.
\newblock {Spatial aggregation of holistically-nested networks for automated
  pancreas segmentation}.
\newblock {\em Lecture Notes in Computer Science (including subseries Lecture
  Notes in Artificial Intelligence and Lecture Notes in Bioinformatics)}, 9901
  LNCS:451--459, 2016.

\bibitem{Roth2017}
H.~R. Roth, H.~Oda, Y.~Hayashi, M.~Oda, N.~Shimizu, M.~Fujiwara, K.~Misawa, and
  K.~Mori.
\newblock Hierarchical 3d fully convolutional networks for multi-organ
  segmentation.
\newblock 2017.

\bibitem{Sabuncu2010}
M.~Sabuncu, B.~Yeo, K.~van Leemput, B.~Fischi, and P.~Golland.
\newblock A generative model for image segmentation based on label fusion.
\newblock {\em IEEE Transactions on Medical Imaging}, 29:1714--1729, 2010.

\bibitem{Saito2016}
A.~Saito, S.~Nawano, and A.~Shimizu.
\newblock {Joint optimization of segmentation and shape prior from
  level-set-based statistical shape model, and its application to the automated
  segmentation of abdominal organs}.
\newblock {\em Medical Image Analysis}, 2105.

\bibitem{Setio2016}
A.~A.~A. Setio, F.~Ciompi, G.~Litjens, P.~Gerke, C.~Jacobs, S.~J. {Van Riel},
  M.~M.~W. Wille, M.~Naqibullah, C.~I. Sanchez, and B.~{Van Ginneken}.
\newblock {Pulmonary Nodule Detection in CT Images: False Positive Reduction
  Using Multi-View Convolutional Networks}.
\newblock {\em IEEE Transactions on Medical Imaging}, 35(5):1160--1169, 2016.

\bibitem{Shelhamer17}
E.~Shelhamer, J.~Long, and T.~Darrell.
\newblock Fully convolutional networks for semantic segmentation.
\newblock {\em IEEE Transactions on Pattern Analysis and Machine Intelligence},
  39(4):640--651, April 2017.

\bibitem{Wang2004}
Z.~Wang, A.~C. Bovik, H.~R. Sheikh, and E.~P. Simoncelli.
\newblock {Image quality assessment: From error visibility to structural
  similarity}.
\newblock {\em IEEE Transcations on Image Processing}, 13(4):600--612, 2004.

\bibitem{Warfield2004}
S.~K. Warfield, K.~H. Zou, and W.~M. Wells.
\newblock {Simultaneous truth and performance level estimation (STAPLE): An
  algorithm for the validation of image segmentation}.
\newblock {\em IEEE Transactions on Medical Imaging}, 23(7):903--921, 2004.

\bibitem{Wolz2013}
R.~Wolz, C.~Chu, K.~Misawa, M.~Fujiwara, K.~Mori, and D.~Rueckert.
\newblock {Automated abdominal multi-organ segmentation with subject-specific
  atlas generation}.
\newblock {\em IEEE Transactions on Medical Imaging}, 32(9):1723--1730, 2013.

\bibitem{Xie2015}
S.~Xie and Z.~Tu.
\newblock Holistically-nested edge detection.
\newblock In {\em {IEEE} International Conference on Computer Vision (ICCV)},
  2015.

\bibitem{Zhou2017}
Y.~Zhou, L.~Xie, W.~Shen, Y.~Wang, E.~K. Fishman, and A.~L. Yuille.
\newblock A fixed-point model for pancreas segmentation in abdominal {CT}
  scans.
\newblock In {\em Medical Image Computing and Computer Assisted Intervention
  (MICCAI)}, 2017.

\bibitem{Zhuang2016}
X.~Zhuang and J.~Shen.
\newblock {Multi-scale patch and multi-modality atlases for whole heart
  segmentation of MRI}.
\newblock {\em Medical Image Analysis}, 31:77--87, 2016.

\bibitem{Zu2017}
C.~Zu, Z.~Wang, D.~Zhang, P.~Liang, Y.~Shi, D.~Shen, and G.~Wu.
\newblock {Robust multi-atlas label propagation by deep sparse representation}.
\newblock {\em Pattern Recognition}, 63:511--517, 2017.

\end{thebibliography}

\bibliographystyle{abbrv}

\end{document}